\documentclass[12pt]{article}
\usepackage{amsmath}
\usepackage{amsfonts}
\usepackage{amssymb}
\usepackage{graphicx}
\usepackage[round]{natbib}% \bibliography
\usepackage{float}
\usepackage{makecell}
\usepackage{multirow}
\usepackage{varwidth}
\usepackage{indentfirst}
\usepackage[table]{xcolor}
\usepackage{hyperref}
\usepackage{algorithm}
\usepackage{algpseudocode}
\usepackage{xfrac}
\usepackage{caption}
\usepackage{subfig}

\usepackage{setspace} 
\usepackage{multicol}
\usepackage[margin=1in]{geometry}
\newcommand{\vecx}{\mathbf{x}}

\newcommand{\matL}{\mathbf{L}}
\newcommand{\matX}{\mathbf{X}}

\newcommand{\veca}{\mathbf{a}}
\newcommand{\vecb}{\mathbf{b}}

\newcommand{\vecX}{\mathbf{X}}
\newcommand{\matI}{\mathbf{I}}
\newcommand{\vecvartheta}{\boldsymbol{\vartheta}}

\newcommand{\matSigma}{\boldsymbol{\Sigma}}

\newcommand{\vecmu}{\boldsymbol{\mu}}

\definecolor{Gray}{gray}{0.7}
\definecolor{lightGray}{gray}{0.9}

\usepackage{tabularx}
\usepackage{makecell}
\usepackage{array}
\usepackage{epsfig}
\usepackage{svg}

\usepackage{subfig}

\title{Turtle Shell Clustering: A Mixture Approach to Discriminative Clustering with Applications to Flow Cytometry and Other Data}
\author{Mackenzie R. Neal*, Paul D. McNicholas*, and Arthur White** }
\date{\small *Department of Mathematics \& Statistics, McMaster University, Ontario, Canada \newline \small **School of Computer Science \& Statistics, Trinity College Dublin, Ireland}

\begin{document}

\maketitle
\begin{abstract}
Generative approaches to clustering provide information on geometric properties of clusters, whereas discriminative approaches provide boundaries between clusters. Ideas from both approaches are incorporated to present a fully unsupervised, probabilistic, and discriminative clustering method via a regularized mutual information objective function, wherein a mixture of mixtures of Gaussian and uniform distributions is used for formulation of the conditional model. Overfitting is avoided by the introduction of a regularizing term and a cluster merge step, similar to those applied in reversible jump Markov chain Monte Carlo methods used in Bayesian clustering. Consequently, the turtle shell method --- a fully unsupervised clustering method capable of estimating non-linear boundary lines, automatically selecting the number of components, and capturing intuitive clusters in the presence of data abnormalities such as noise and/or irregular cluster shapes --- is introduced. We test this method on various simulated and real datasets commonly explored in clustering research, and extend the analysis to datasets arising from flow cytometry experiments and image analysis. 

\end{abstract}

\section{Introduction}     
Clustering is an unsupervised learning problem in which the goal is to uncover groupings of similar observations; the definition of `similar' depends on the problem at hand and the method used to solve it. Generative approaches assume that `similar' datapoints arise from the same probabilistic distribution and define clusters in terms of their geometric properties~\citep{mcnicholas16a,mcnicholas16b}. Additionally, as generative clustering methods possess a probabilistic foundation, information on cluster membership and uncertainty is provided. Discriminative approaches assume that `similar' datapoints are datapoints that are not well-separated from each other but are well-separated from datapoints belonging to other clusters~\citep{bridle1991unsupervised,grandvalet2004semi,krause2010discriminative}. These approaches are sometimes preferred because they require fewer assumptions about the data and provide intuitive clusters. However, as these methods lack a probabilistic foundation, they are unsuitable for problems that require knowledge of classification uncertainty. Further, the lack of a probabilistic foundation reduces interpretability and often makes decisions pertaining to the number of components more difficult.  

We propose an approach to bridge these gaps, bringing together generative approaches and discriminative approaches by combining the work of \citet{krause2010discriminative} on unsupervised discriminative clustering with that of \citet{browne2011model} on robust mixture modeling, and \citet{zhang2003algorithms} on selection of the number of clusters in the Bayesian clustering paradigm. The result is the turtle shell method, a method where classification uncertainty is easily calculated, geometric properties of clusters are provided, the number of components is selected automatically, non-linear boundaries are drawn, and well-separated clusters are obtained. This approach may be applied to any problem/dataset with these goals in mind, as evidenced in Section \ref{results}, where we implement the turtle shell on various benchmark clustering datasets and image datasets. Further, this work is well-suited for analyzing cellular data arising from flow cytometry experiments in immunological research. Immunological researchers are often interested in identifying sub-populations of cells within samples to gain an improved understanding of the disease impact on cell types. However, traditional methods used for carrying out such analyses are often manual and subjective, with the possibility of different researchers arriving at different results on the same dataset as boundary lines are manually drawn by researchers \citep{aghaeepour2013critical}. Making automated clustering methods essential in flow cytometry, especially discriminated methods --- a currently underdeveloped area. In Section \ref{results}, we apply the proposed method to two benchmark flow cytometry datasets and compare to two prominent automated clustering methods designed for flow cytometry: cytometree~\citep{commenges2018cytometree} and PhenoGraph~\citep{levine2015data}. 

The paper is outlined as follows: In Section \ref{back} we discuss motivating approaches for the work proposed herein and traditional clustering methods for flow cytometry datasets. In Section \ref{method}, we present our turtle shell method including the estimation and the component selection procedures. In Section \ref{results}, we examine the performance of the turtle shell on both simulated and real data. Finally, in Section \ref{discuss}, we review the results and discuss recommendations for future work.

\section{Background} \label{back}
\subsection{Model-Based Clustering}
One of the most common generative clustering methods is model-based clustering, wherein the statistical model emerges from the assumption that the data, $\matX = (\vecx_1,\ldots,\vecx_N)$ where $\vecx_i = (x_{i1},\ldots,x_{iD})'$, arises from a finite mixture distribution \citep{mcnicholas16a}. Commonly, the component densities are taken to be Gaussian, resulting in the mixture density
$$f(\vecx_i | \vecvartheta) =  \sum_{k=1}^K \tau_k \phi (\vecx_i | \vecmu_k, \matSigma_k),$$
where $K$ is the number of clusters or the order of the mixture, $\tau_k$ is the mixing proportion for cluster $k$, $\vecmu_k$ is the mean vector for cluster $k$, and $\matSigma_k$ is the covariance matrix of cluster $k$. Let $y_i \in \{1,\ldots, K\}$ be the cluster membership of observation $i$, where $i=1,\ldots,N$. The corresponding \textit{a~posteriori} probability is 
$$p(y_{i}=k | \vecx_i) = \frac{\tau_k \phi (\vecx_i | \vecmu_k, \matSigma_k)}{\sum_{g=1}^K \tau_g \phi (\vecx_i | \vecmu_g, \matSigma_g)},$$
where $y_{i}$ follows a categorical distribution on $K$ categories with probabilities defined by $\tau_1,\ldots,\tau_K$. From this point onwards we denote $y_i=k$ as $y_{ik}$ for simplicity. Once all parameters are estimated --- often through maximum likelihood estimation --- the predicted classifications can be calculated. For convenience we write this as 
$$\hat{y}_{ik}= \frac{\hat{\tau}_k \phi (\vecx_i | \hat{\vecmu}_k, \hat{\matSigma}_k)}{\sum_{g=1}^K \hat{\tau}_g \phi (\vecx_i | \hat{\vecmu}_g, \hat{\matSigma}_g)}.$$
The predicted classifications are soft, meaning each datapoint has a probability of belonging to each cluster and therefore classification uncertainty can be quantified \citep{mcnicholas16a}.

With maximum likelihood estimation, various functions of the likelihood arise as possible criteria for selecting the number of components, $K$. The most common order selection criterion is the Bayesian information criterion [BIC; \citep{schwarz78}], $\text{BIC} = 2l(\hat{\vecvartheta}) - \rho \log N,$ where $l(\hat{\vecvartheta})$ is the maximized log-likelihood, $\rho$ is the number of free parameters, and $N$ is the number of observations. Alternatively, an entropy penalty is added to the BIC for the integrated complete-data likelihood [ICL; \citep{biernacki2002assessing}], which can be approximated via 
\begin{equation*}
   \text{ICL} \approx \text{BIC} + 2 \sum_{i = 1}^n \sum_{k = 1}^K \text{MAP}\{\hat{y}_{ik}\} \log \hat{y}_{ik}, 
\end{equation*}
where  $\text{MAP}\{\hat{y}_{ik}\}$ is the maximum \textit{a~posteriori} classification, i.e.,
\begin{equation*}
    \text{MAP}\{\hat{y}_{ik}\} = \begin{cases} 
     1 & \text{if $k = \arg \max_g  \{\hat{y}_{ig}\}$},\\
     0 & \text{otherwise.}
   \end{cases}
\end{equation*}

\begin{figure}[ht]% 
    \subfloat[\centering Estimated clusters by the EM with the BIC.]{{\includegraphics[width=0.47\linewidth]{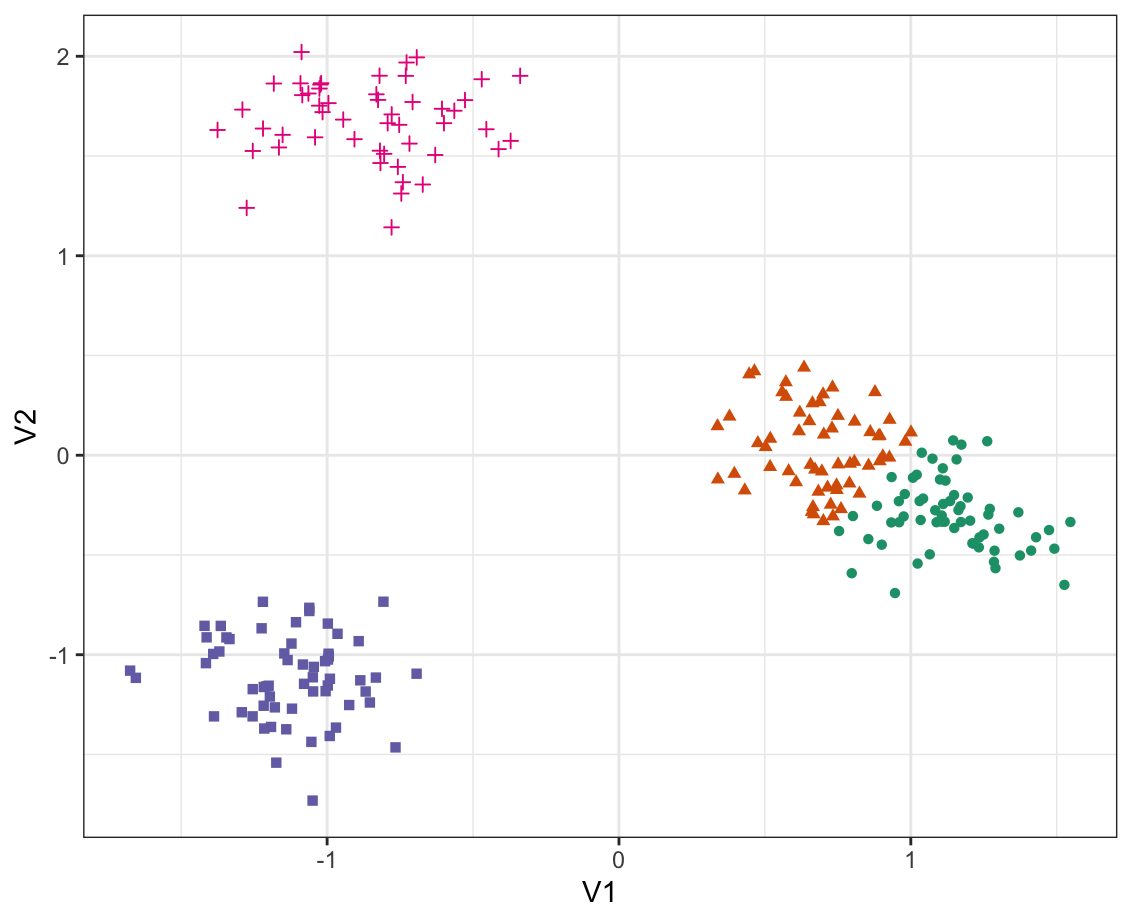}}}%
    \qquad
    \subfloat[\centering Estimated clusters by the EM with the ICL.]{{\includegraphics[width=0.47\linewidth]{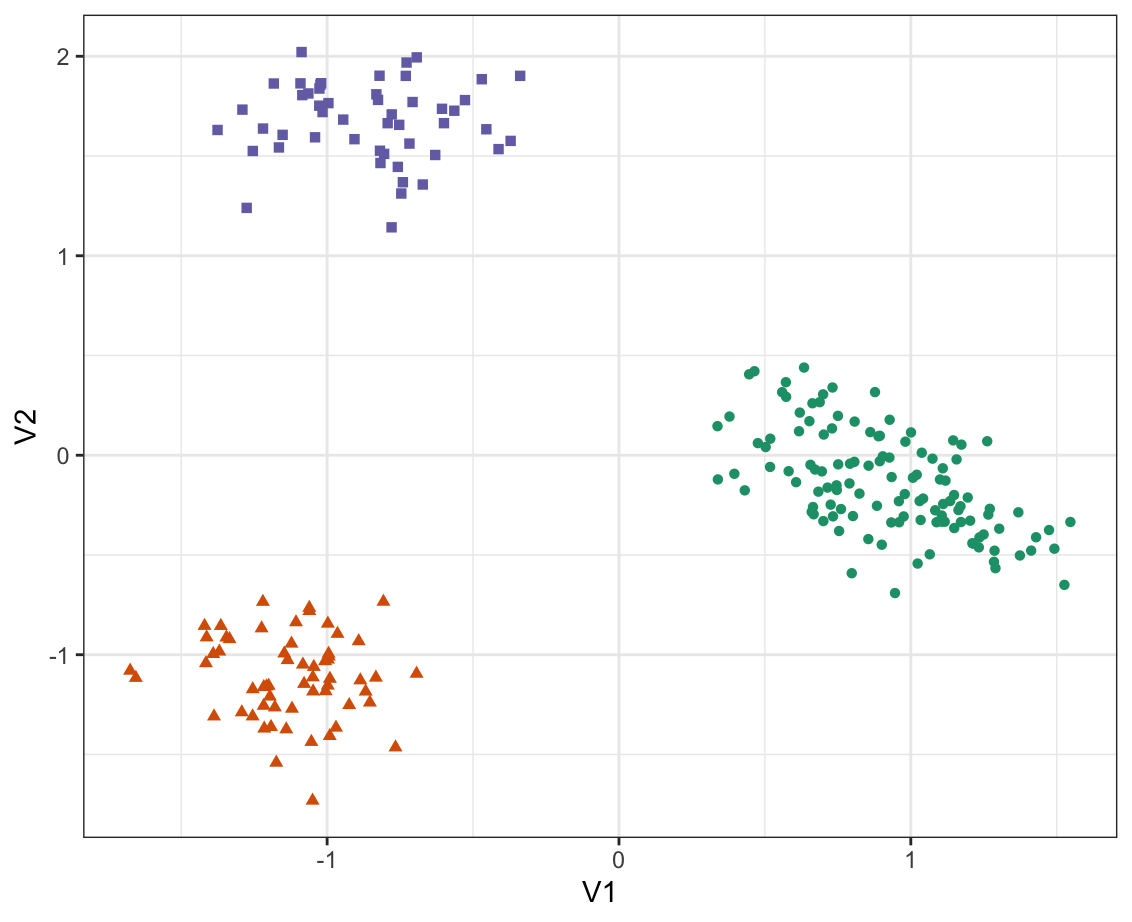}}}%
    \caption{Clusters from the EM estimation of a GMM when (a) the BIC is used to select $K$ and when (b) the ICL is used to select $K$.}
    \label{bic_ex}
\end{figure}

Many model-based clustering methods rely on the BIC to perform order selection. There are instances, however, when the BIC may separate similar observations into distinct clusters. For example, when component densities are assumed to be Gaussian, multiple clusters that adhere to Gaussian properties may be obtained, but may be less intuitive than a single merged cluster. This behaviour is exhibited in Figure \ref{bic_ex} wherein data from a four-component Gaussian mixture model (GMM) is randomly generated and a GMM is estimated via the expectation-maximization (EM) algorithm \citep{dempster77} for $K= 2, \ldots, 9$ using both the BIC and the ICL for order selection. We see that the BIC selects four clusters, corresponding to the four data-generating Gaussian distributions. However, the ICL selects the more intuitive three-cluster solution; see \citet{baudry15} for an in-depth examination of this behaviour. 

\begin{figure}[ht]
    \centering
    \includegraphics[scale=0.4]{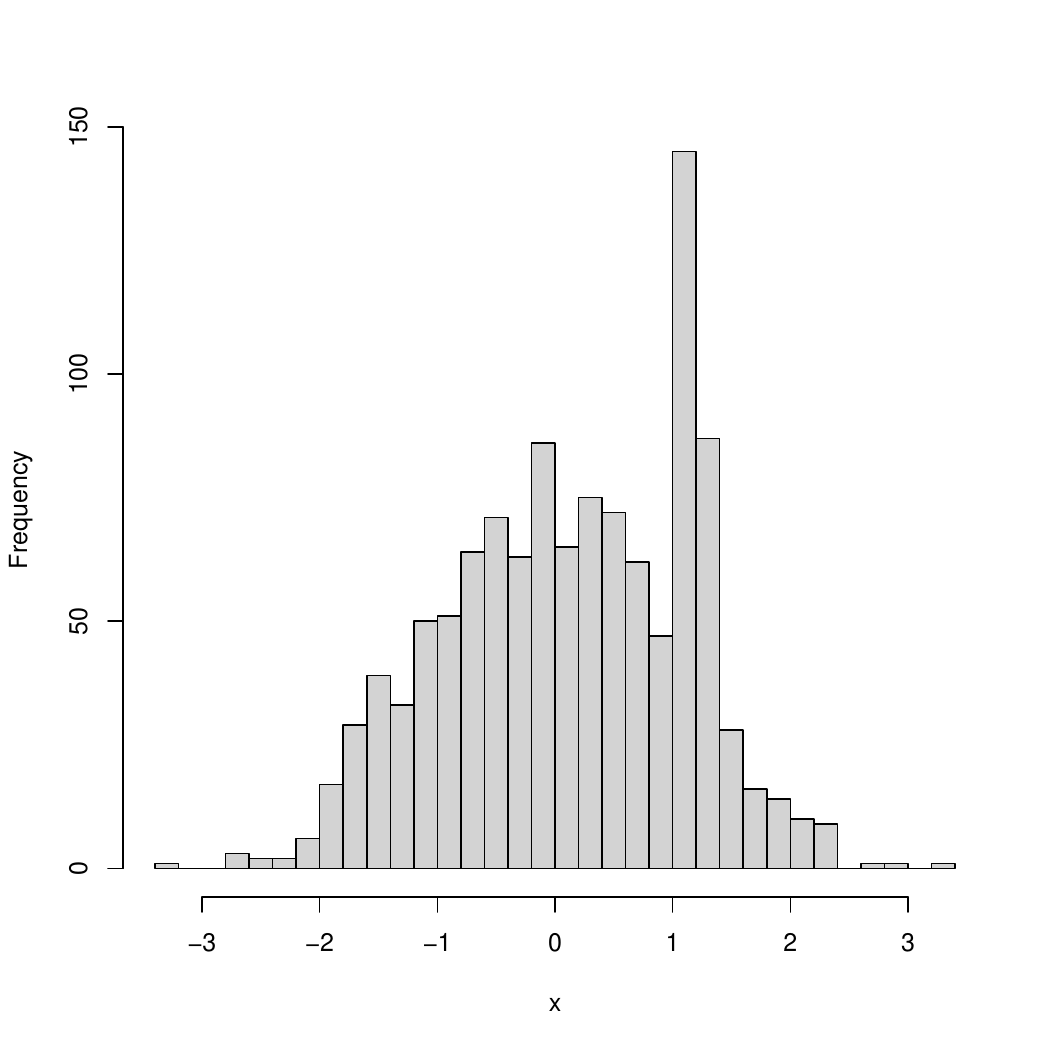}
    \caption{Histogram of data generated from a mixture of Gaussian and uniform distributions.}
    \label{histGU}
\end{figure}

So far, all discussions on model-based clustering have been within the Gaussian paradigm; however, in practice, it is often unrealistic to make this assumption. As a result, much work has been done to extend model-based clustering to non-normal clusters~\citep{mcnicholas16b}. The mixture of mixtures of Gaussian and uniform distributions model is an effective solution to this challenge~\citep{browne2011model}. The component density of this mixture of mixtures is
$$f(\vecx_i | \vecvartheta_k) =  \omega_k\phi (\vecx_i | \vecmu_k, \matSigma_k) + (1-\omega_k)u(\vecx_i | \boldsymbol{\zeta}_k),$$
where $\omega_k \in (0,1)$ is the inner mixing proportion and $u(\vecx_i | \boldsymbol{\zeta}_k)$ is the multivariate uniform distribution with the range for each variable denoted by the elements of the $2D$-dimensional vector $\boldsymbol{\zeta}_k$ \citep{browne2011model}. The mixture of mixtures allows for capturing Gaussian clusters wherein a burst of data beyond what is expected under the Gaussian is observed in the cluster, as visualized in Figure \ref{histGU}. We extend the work of \citet{browne2011model} herein, detailed in Section \ref{GU_spec}, to control for noise and high-density regions. 

\subsection{Maximization of Mutual Information}
Discriminative clustering methods estimate the boundary lines between clusters. \citet{krause2010discriminative} introduced the discriminative clustering method called Regularized Information Maximization, or RIM, a probabilistic method that extends the work of \citet{bridle1991unsupervised} and \citet{grandvalet2004semi}. In RIM, a regularized mutual information objective is implemented as 
\begin{align*}
    F(\vecvartheta; \vecX,\lambda) & = I_{\vecvartheta}\{y;\vecx\} - R(\vecvartheta;\lambda)\\ 
    &=H\{\hat{p}(y;\vecvartheta)\} - \frac{1}{N}\sum_i H\{p(y_i|\vecx_i,\vecvartheta)\} - R(\vecvartheta;\lambda)
\end{align*} 
where $\hat{p}(y;\vecvartheta) = \frac{1}{N}\sum_{i=1}^N p(y_i|\vecx_i;\vecvartheta)$; $H\{\hat{p}(y;\vecvartheta)\} - \frac{1}{N}\sum_i H\{p(y_i|\vecx_i,\vecvartheta)\}$ is the mutual information between the unlabelled data $\vecx_i$ and the unobserved class labels $y_i$, and $R(\vecvartheta;\lambda)$ is a regularizing term for order selection. The conditional entropy term, $- \frac{1}{N}\sum_i H\{p(y_i|\vecx_i,\vecvartheta)\}$, ensures boundary lines don't overlay data points; however, conditional entropy may be reduced by the removal of good boundaries. To counteract this, the entropy of class labels, $H\{\hat{p}(y;\vecvartheta)$ is included in the objective function, as this term is maximized under class balance. Together these terms make up mutual information (MI), $\mbox{MI} = H\{\hat{p}(y;\vecvartheta)\} - \frac{1}{N}\sum_i H\{p(y_i|\vecx_i,\vecvartheta)\}$. A regularizing term, $R(\vecvartheta;\lambda)$, is needed to avoid overfitting, as MI is trivially maximized when every data point is assigned to its own cluster --- for alternative solutions to this problem, see \cite{ohl2022generalised}. In maximizing $F(\vecvartheta; \vecX,\lambda)$ using some quasi-Newton method, such as L-BFGS-B,  the probabilistic classifier $p(y_i | \vecx_i)$ is trained.

\begin{figure}[ht]
    \centering
        \includegraphics[width=0.4\linewidth]{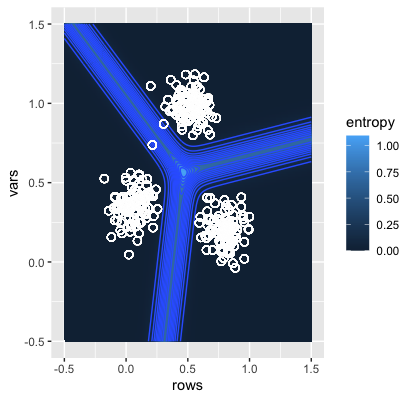}
      \caption{Example of RIM clustering results when a multi-logit is assumed.}
        \label{fig:multi_logit}
    \end{figure}

\citet{krause2010discriminative} construct this objective function using a multi-logit regression model, with a regularizing term applied to the set of weights from the regression model. In Figure \ref{fig:multi_logit}, we see the resulting partition from this model on a simple clustering problem. 

The multi-logit model has limitations, primarily the inability to capture non-linear boundary lines. \citet{krause2010discriminative} discuss the use of kernels to overcome this limitation; however, due to many datasets containing areas of both high density and high sparsity, kernels would be expected to perform poorly \citep{marin2017kernel}. We look to model-based clustering to solve these problems, combining ideas from RIM and the mixture of mixtures of Gaussian and uniform distributions, detailed in Section \ref{method}. 

\subsection{Clustering Flow Cytometry Data: Manual Gating}
Automated clustering methods for flow cytometry are an ongoing area of intensive research \citep{qian2010elucidation, hejblum2019sequential, van2015flowsom, levine2015data,Doherty2025}, but discriminative approaches are under-explored in this setting. However, such approaches correspond well to the traditionally used manual gating methods, in which the practitioner gates/clusters cells by manually drawing boundary lines for one variable at a time until pleased with the resulting partition \citep{aghaeepour2013critical}. That is, manual gating emphasizes the identification of distinct populations via discriminative approaches. 

It is evident that automated clustering methods are needed as the manual gating procedure is subjective, irreproducible, and highly influenced by variable order~\citep{aghaeepour2013critical,aghaeepour2016benchmark,Liu2019comparison}. If these methods are to mimic the results obtained from manual gating, they must be able to handle non-normal clusters, as there is no guarantee of normality, or even convexity,  from the manual gating procedure. Additionally, as data arising from flow cytometry technology are on the cellular level, many similar observations are observed; thus, any effective automated clustering method must control for areas of high density and must select the number of clusters in a computationally effective way. The proposed turtle shell method, detailed in Section \ref{method}, overcomes both of these data issues, making it well-suited for flow cytometry. 

\section{Methodology} \label{method}
\subsection{General Framework}
The objective function from \citet{krause2010discriminative} simplifies to,

\begin{align*}
    F(\vecvartheta; \vecX,\lambda) & = I_{\vecvartheta}\{y;\vecx\} - R_1(\vecvartheta;\lambda_1)\\ 
    &=H\{\hat{p}(y;\vecvartheta)\} - \frac{1}{N}\sum_i H\{p(y_i|\vecx_i,\vecvartheta)\} - R_1(\vecvartheta;\lambda_1)\\
    &= \frac{1}{N} \sum_{i=1}^N \sum_{k=1}^K p(y_{ik}|\vecx_i;\vecvartheta_k) \text{log}\frac{p(y_{ik}|\vecx_i;\vecvartheta_k)}{\hat{p}(y_k;\vecvartheta_k)} - R_1(\vecvartheta;\lambda_1).
\end{align*} 
We assume $\vecX$ arises from a finite mixture model; thus the conditional model, $p(y_{ik}|\vecx_i)$ becomes the \textit{a~posteriori} probability from a mixture model, that is
\begin{align*}
p(y_{ik}|\vecx_i) &= \frac{s(\pi_k) f_k(\vecx_i; \vecvartheta_k)}{\sum_{g=1}^K s(\pi_g) f_g(\vecx_i; \vecvartheta_g)},
\end{align*}

where the mixing proportion $\tau_k$ is replaced by the soft-max transformation of $\pi_k$ to ensure the properties of the mixing proportions remain intact, where $\pi_k \in \mathbb{R}$, that is $\tau_k = s(\pi_k) = \frac{exp(\pi_k)}{\sum_{g=1}^K exp(\pi_g)}$, and $f_k(\vecx_i;\vecvartheta_k)$ is the $k^{\text{th}}$ component density, specified in Section \ref{GU_spec}. A natural choice for the regularizing term, $R_1(\vecvartheta;\lambda_1)$, is to impose a penalty on the mixing proportions to avoid overfitting and cluster instability. As with hidden Markov models \cite{hung2013hidden, zou2024bayesian}, we select the regularizing term to be
\begin{align*}
R_1(\vecvartheta;\lambda_1)&=-\lambda_1 \sum_{k=1}^K \log{s(\pi_k)},
\end{align*}
where $\lambda_1 \geq 0 $ is a tuning parameter. Further, small clusters are removed from the model if the following criterion is not met
\begin{align*}
\frac{n_k}{N}=\frac{\sum_{i=1}^N \text{MAP}\{p(y_{ik}|\vecx_i;\vecvartheta_k)\}}{N} \geq a,
\end{align*}
where $\sfrac{n_k}{N}$ corresponds to the proportion of observations in cluster $k$ and $a$ is a predetermined threshold, i.e, 0.01 or $\sfrac{10}{N}$. Small clusters are removed one at a time and the model is re-estimated with each removal to allow the parameters from the retained clusters to adjust in response to the removal. Following \citet{krause2010discriminative}, we first estimate the model with $\lambda_1=0$, using this result as the initialization for the model with $\lambda_1 > 0$. From there, we select the result that obtains the best average silhouette width (ASW), or approximate ASW for high-dimensional datasets as implemented in the \textsf{R} package \texttt{bluster} \citep{bluster}.

Similar to \citet{krause2010discriminative}, we estimate the model by maximization of $F(\vecvartheta; \vecX,\lambda)$ using a quasi-Newton algorithm. Before doing so, the conditional model, $p(y_{ik}|\vecx_i;\vecvartheta_k)$, must be fully specified by selection of the component densities, $f_k(\vecx_i;\vecvartheta_k)$, and the gradients must be derived. With our clustering goals in mind, and considering motivating data such as flow cytometry data, we assume $\vecX$ arises from a mixture of mixtures of Gaussian and uniform distributions, we refer to this as the full turtle shell model, detailed in Section \ref{GU_spec}. To further motivate the full turtle shell, comparisons to a simple turtle shell, whereby we assume $\vecX$ arises from a mixture of Gaussian distributions, are included in Sections \ref{init_details} and \ref{results}. 

\subsection{Full Turtle Shell Specifications}\label{GU_spec}
To account for potential high-density areas, noise, and skewness and to obtain non-linear boundaries, we assume $\vecX$ belongs to  mixture of mixtures of Gaussian and uniform distributions resulting in the conditional model 
$$p(y_{ik}|\vecx_i; \vecvartheta_k) = \frac{s(\pi_k) [\omega_k\mathcal{N}(\vecmu_k, [\matL_k\matL_k^\top]^{-1}) + (1- \omega_k)U(\veca_k,\vecb_k)]}{\sum_{g=1}^K s(\pi_g) [\omega_g\mathcal{N}(\vecmu_g, [\matL_g\matL_g^\top]^{-1}) + (1- \omega_g)U(\veca_g,\vecb_g)]},$$
where $\matL_k$ is the lower triangular matrix from the Cholesky decomposition of the precision matrix of cluster $k$, and $\veca_k$ and $\vecb_k$ are the $D$-dimensional bounds for the uniform component of cluster $k$, where $\veca_k = (a_{1k},\ldots,a_{Dk}),\vecb_k = (b_{1k},\ldots,b_{Dk}), \text{ with } a_{ik} < b_{ik} \quad \forall \quad 1\leq i \leq D \text{ and } 1\leq k \leq K$. 

This specification of the conditional model introduces the potential problem of the inner components --- the Gaussian and the uniform distributions of a given cluster --- moving away from each other during the estimation procedure. In doing so, an underestimation of the number of clusters may occur if what should be two separate clusters is estimated as one cluster with distant inner components. To overcome this problem, we introduce a second penalty term to the objective function, wherein the distance between the mean of the Gaussian component and the mean of the uniform component is penalized, taking into consideration the shape of the cluster, as 

$$R_2(\vecvartheta;\lambda_2)=\lambda_2 \sum_{k=1}^K \bigg( \frac{\vecb_k+\veca_k}{2} - \vecmu_k \bigg)^\top (\matL_k\matL_k^\top) \bigg( \frac{\vecb_k+\veca_k}{2} - \vecmu_k \bigg),$$
resulting in the new objective function
\begin{align*}
    F(\vecvartheta; \vecX,\lambda) & = I_{\vecvartheta}\{y;\vecx\} - R_1(\vecvartheta;\lambda_1) -R_2(\vecvartheta;\lambda_2).
\end{align*} 
With $R_2(\vecvartheta;\lambda_2)$ a new tuning parameter, $\lambda_2$, is introduced. Selection of both tuning parameters, $\lambda_1$ and $\lambda_2$, is parallelized to reduce computational time. Again, we first estimate the model with $\lambda_1=\lambda_2=0$ and use this result as the initialization to the model with nonzero tuning parameters, similar to \citep{krause2010discriminative}, selecting the final result based on the best ASW. The gradients for $R_1(\vecvartheta;\lambda_1)$ and $R_2(\vecvartheta;\lambda_2)$ are easily obtained and subtracted from the gradients of $I_{\vecvartheta}\{y;\vecx\}$, see Appendix \ref{appendix:appA}. For model estimation, we use the box-constrained quasi-Newton optimization algorithm L-BFGS-B \citep{byrd1995limited}, implemented in the \textsf{R} function \texttt{optim} \citep{R2025}.

\subsubsection{Order Selection}
In many cases, we find additional cluster removal to be necessary, for instance when high-density regions are divided into multiple small clusters during the initialization procedure,  discussed in Section \ref{init_details}, or when some local optimum has been achieved. To overcome this, we implement a merge step analogous to those implemented in Bayesian estimation of mixture models via reversible jump Markov chain Monte Carlo methods [RJMCMC; \citep{zhang2004learning}]. To determine which clusters to merge first, we calculate the following entropy term for each cluster
\begin{align*}
\sum_{i=1}^N p(y_{ik}|\vecx_i;\vecvartheta_k) \log{p(y_{ik}|\vecx_i;\vecvartheta_k)},
\end{align*}
and obtain the cluster with the highest uncertainty. We then determine the most similar cluster, analogous to \citet{zhang2003algorithms}, where similarity is defined as
\begin{align*}
J_{\text{merge}}(i,k;\vecvartheta) &= \frac{P_i(\vecvartheta)'P_k(\vecvartheta)}{||P_i(\vecvartheta)|| ||P_k(\vecvartheta)||},
\end{align*} 
where $P_i(\vecvartheta)$ is the $N$-dimensional vector of \textit{a~posteriori} probabilities for the $i$th component.  In \citet{zhang2003algorithms} a mixture of Gaussian distributions is considered and the merged component relates to the two clusters under consideration via
\begin{align*}
    &s(\pi_{i'}) = s(\pi_i)+s(\pi_k),\\ 
    &\vecmu_{i'} = \frac{s(\pi_i)\vecmu_i+s(\pi_k)\vecmu_k}{s(\pi_{i'})},\\
    &\matSigma_{i'} = \frac{1}{s(\pi_{i'})}\bigg(s(\pi_i)[\matSigma_i+\vecmu_i\vecmu_i']+s(\pi_k)[\matSigma_k+\vecmu_k\vecmu_k']\bigg) - \vecmu_{i'}\vecmu_{i'}'.
\end{align*} 
We extend this to the mixture of mixtures of Gaussian and uniform distributions with the addition of
\begin{align*}
&\omega_{i'} = \frac{s(\pi_i)\omega_i+s(\pi_k)\omega_k}{s(\pi_{i'})},\\
 &\veca_{i'} = \frac{s(\pi_i)\veca_i+s(\pi_k)\veca_k}{s(\pi_{i'})},\\
  &\vecb_{i'} = \frac{s(\pi_i)\vecb_i+s(\pi_k)\vecb_k}{s(\pi_{i'})}.\\
\end{align*} 

Once the parameters of the merged cluster are obtained, we re-estimate the \textit{a~posteriori} probabilities, obtain new maximum \textit{a~posteriori} classifications, and recalculate the ASW. If the ASW improves from the merge, the merge is accepted, and the model is re-estimated. If the merge is not accepted, the next most uncertain cluster is tested for merging, and this process iterates until either a merge is accepted or all clusters have been tested. The model is only re-estimated once a merge is accepted, making this an efficient procedure for cluster removal.

\subsection{Initialization}\label{init_details}
We study the effect of three initialization procedures on the turtle shell on 200 simulated datasets generated from a six-component mixture of mixtures of Gaussian and uniform distributions, each having 1150 observations and two variables. The first initialization scheme uses $k$-means to initialize the Gaussian means; thus, the initial number of clusters, $k$, must be chosen \textit{a~priori}; as such, we test $k=10,20, \text{ and } 50$. The second initialization scheme implements Louvain community detection \citep{blondel2008fast} on a $k$-nearest neighbours ($k$NN) graph, using the \textsf{R} package \texttt{igraph} \citep{igraph_pack}, to select the initial number of clusters and their Gaussian means. For this initialization scheme, the number of neighbours to consider in the $k$NN graph must be provided \textit{a~priori}; we test $k=15,25, \text{ and } 45$. Lastly, we use Latin hypercube sampling (LHS) as implemented in the \textsf{R} package \texttt{lhs} \citep{lhs_package}, to obtain $k$ evenly spaced nodes in the input space; these nodes serve as the initial Gaussian means. We consider $k=10,20, \text{ and } 50$. We compare the effects of each initialization scheme on the full turtle shell to the effects on a simple turtle shell. These results are visualized in Figure \ref{fig:init}, where the final estimated number of clusters from each method for each initialization scheme and initial parameter value is recorded.

\begin{figure}[H]
  \centering
    \setcounter{subfigure}{0}
    \subfloat[$k$NN and Louvain community detection]{%
  \includegraphics[width=\textwidth]{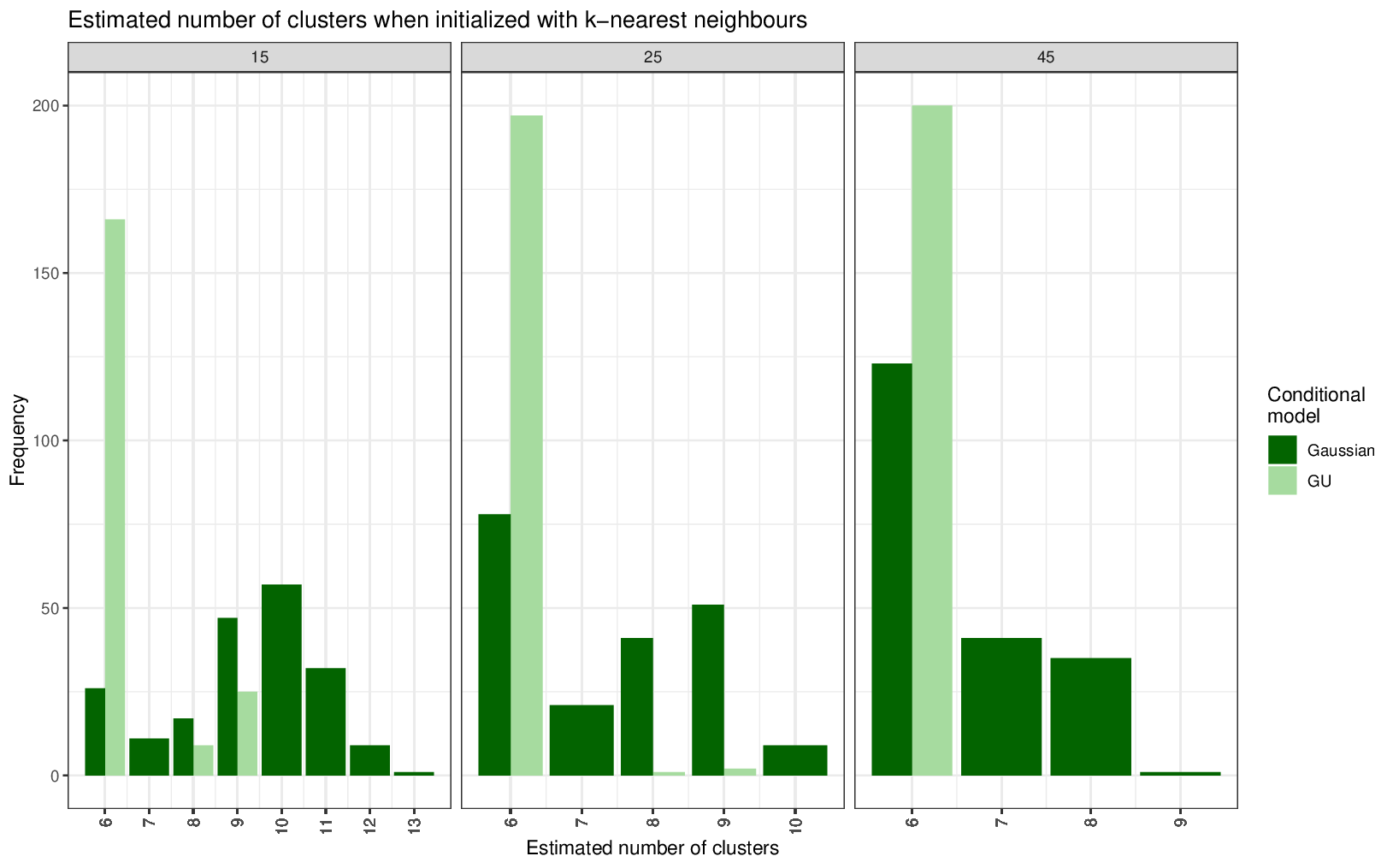}%
}

  \subfloat[LHS]{%
  \includegraphics[width=\textwidth]{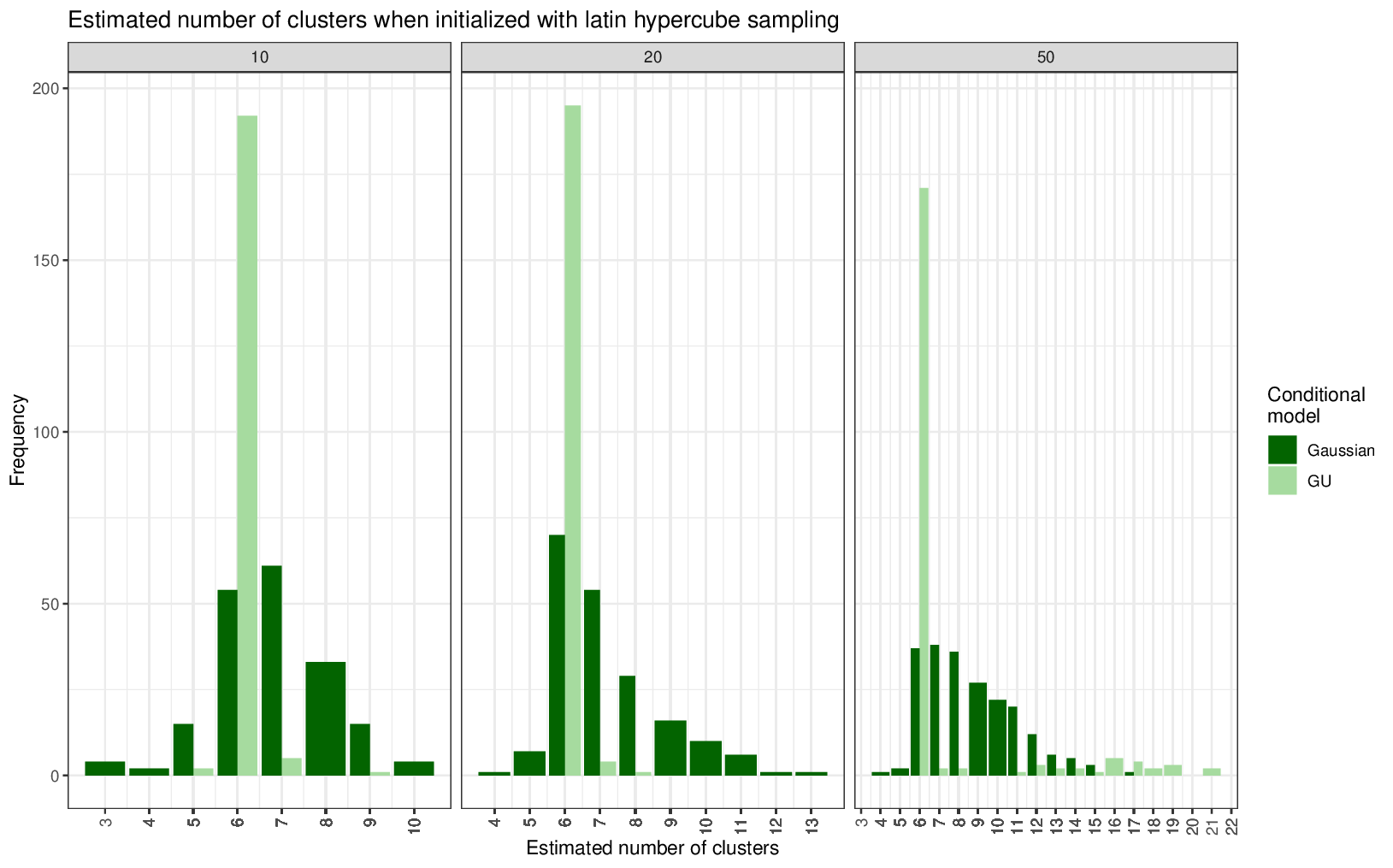}%
}
  \caption{Estimated number of clusters for each initialization method.}
  \label{fig:init}
\end{figure}
%\clearpage
\begin{figure}[ht]
  \ContinuedFloat 
  \centering
\subfloat[$k$-means]{%
  \includegraphics[width=\textwidth]{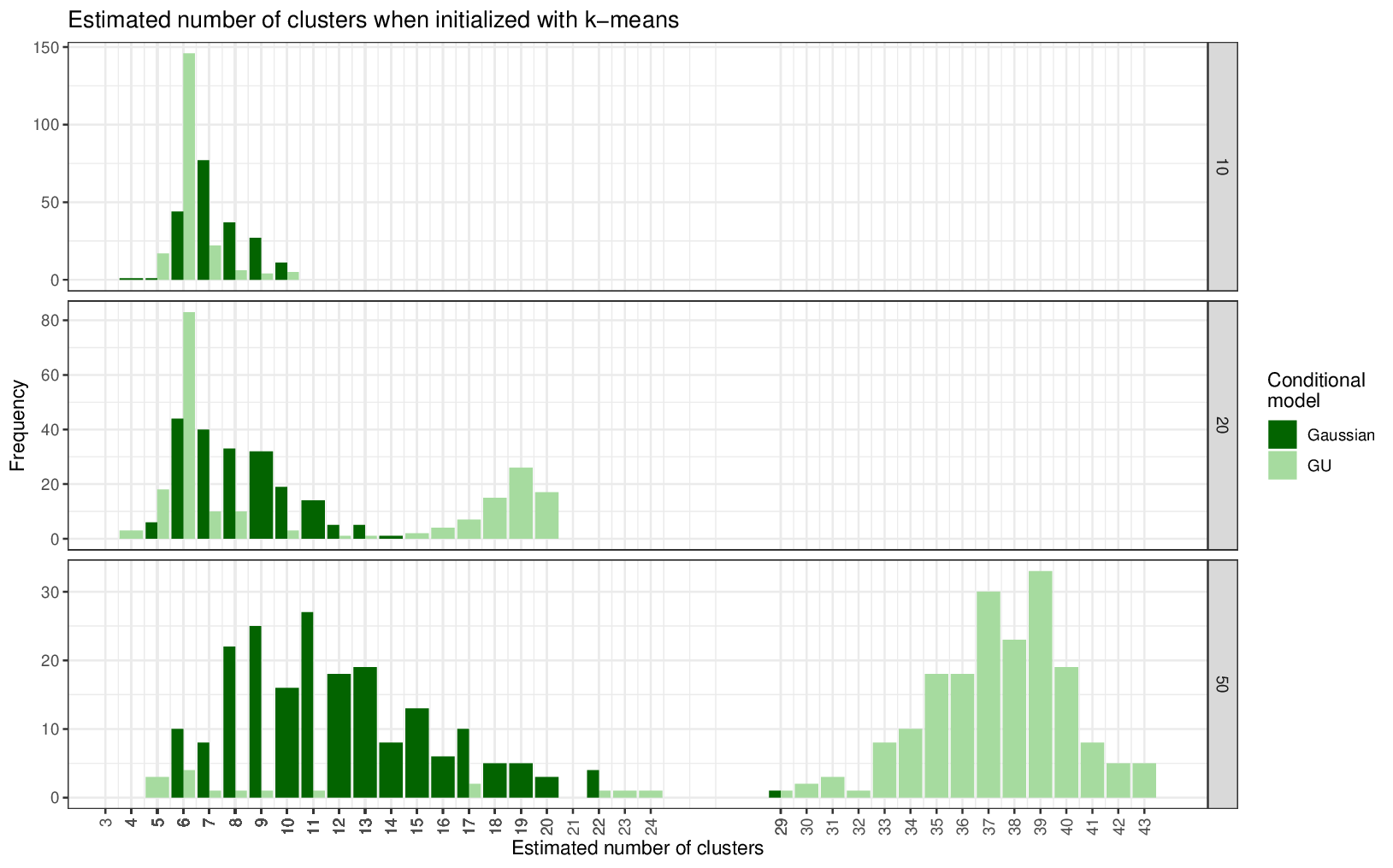}%
}
\caption{Estimated number of clusters for each initialization method.}
\end{figure}
 For the LHS and graphing initializations, the full turtle shell frequently selects the correct number of clusters, regardless of the initial parameter value. Nonetheless, going forward, we use the graphing initialization for computational efficiency, as Louvain community detection will provide an intelligent guess for the initial number of clusters, rather than using a completely arbitrary number as with LHS and $k$-means. That being said, in most instances, the graphing initialization will still overestimate the number of clusters. In this simulation, for example, the initial number of clusters was typically in the range of [9,15]. 
For full details of the chosen initialization procedure, see Algorithm \ref{alg:cap}.

\begin{algorithm}[htbp]
\caption{ Graph partition}\label{alg:cap}
\begin{algorithmic}
 \For{\texttt{i in 1:Nstarts}}
        \State Construct a symmetric, sparse adjacency matrix from a $k$NN graph.
        \State Apply Louvain community detection.
     \State Set $K$ to be the number of communities obtained from Louvain community detection.
     \State Set $\mu_j$ to be the mean of community $j$ $\forall j \in \{1,\ldots,K\}$.
     \State Set $\matSigma_j = (1-\alpha)\matI_D + \alpha \text{Cov}(X_j)$  $\forall j \in \{1,\ldots,K\}$ where $\alpha =  [(i-1)/(\text{Nstarts})]$.
     \State Set $\pi_j = \sfrac{1}{K}$ and $\omega_j = 0.7$ $\forall j \in \{1,\ldots,K\}$.
\State For all $j \in \{1,\ldots,K\}$, randomly choose the location for the $j^{\text{th}}$ uniform component, provided $\mu_j$ lies within it.
\EndFor
\State Calculate the \textit{a~posteriori} probabilities and select the initialization that maximizes the ASW. 
\end{algorithmic}
\end{algorithm} 

\subsection{Toy Example}

\begin{figure}[ht!]% 
  \centering
    \qquad
           \subfloat[\centering  Initial clustering from graphing initials.]{{\includegraphics[width=0.43\linewidth]{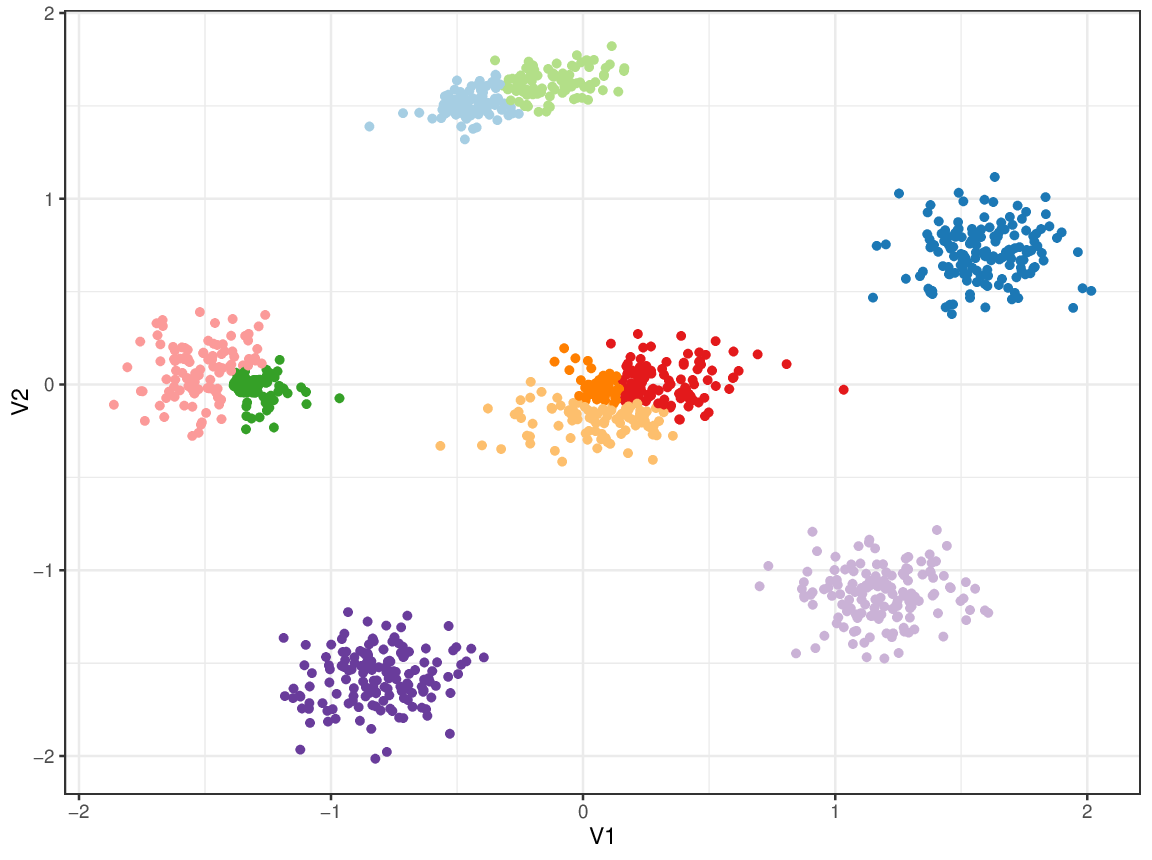}}\label{fig:subfigA}}%
                 \subfloat[\centering  Objective during the model fitting procedure.]{{\includegraphics[width=0.4\linewidth]{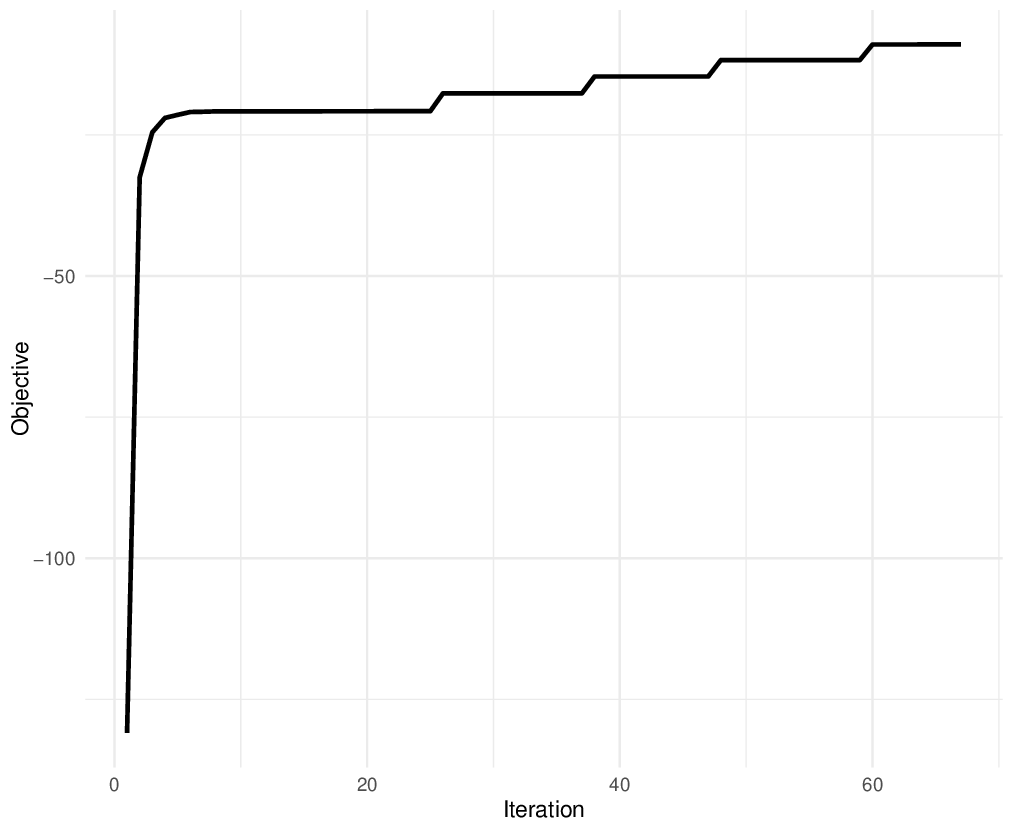}}\label{fig:subfigB}}%

  \centering
    \qquad
        \subfloat[\centering  Estimated partition.]{{\includegraphics[width=0.43\linewidth]{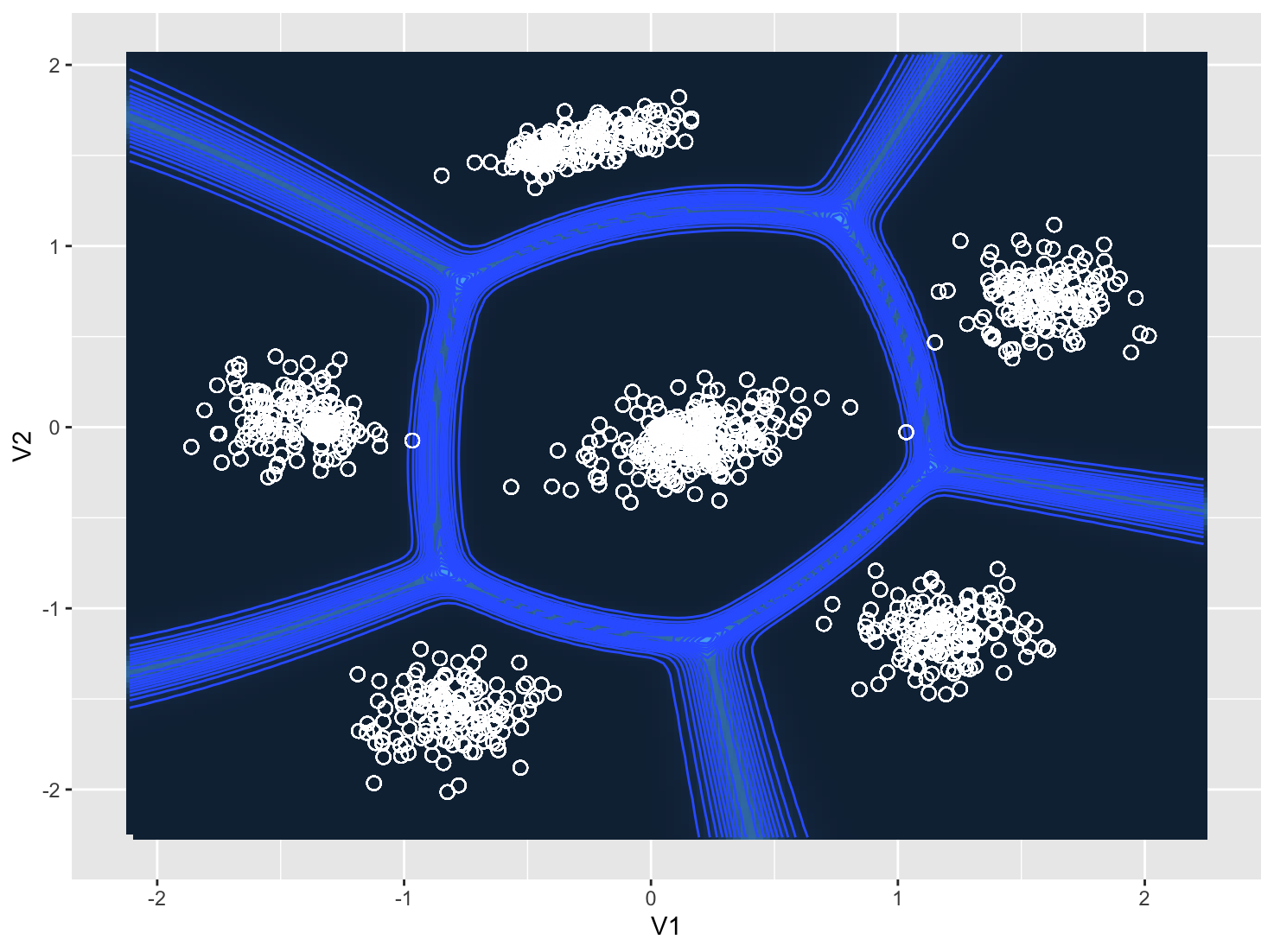}}\label{fig:subfigC}}%
     \subfloat[\centering Final estimated clusters.]{{\includegraphics[width=0.43\linewidth]{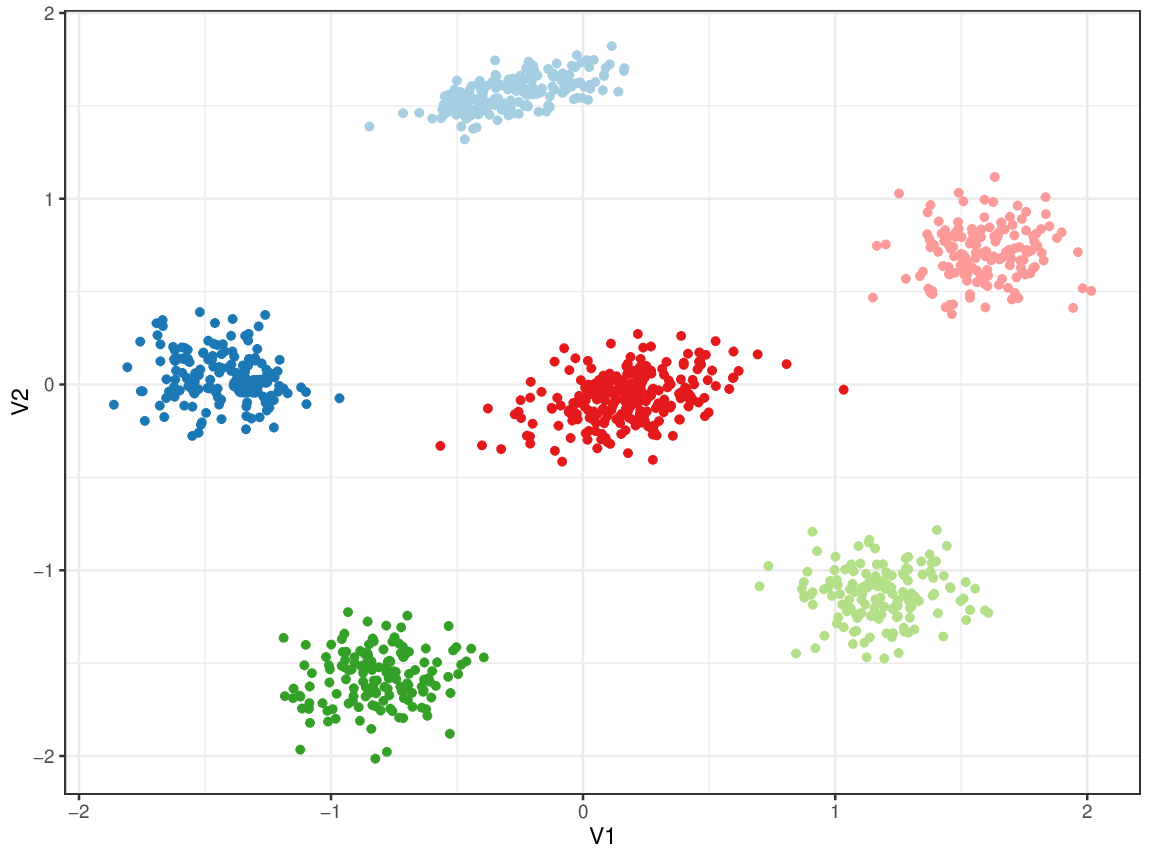}}\label{fig:subfigD}}%
         \caption{An example result on a simulated dataset from Section \ref{init_details}.}
    \label{ex:toy}
\end{figure}

For illustration of how the turtle shell works, including the initialization procedure and the influence of the merge step, we simulate the same dataset as in Section \ref{init_details} once, implement the full turtle shell with $k=25$, and summarize the results in Figure \ref{ex:toy}. Figure \ref{fig:subfigA} provides an illustration of the estimated classes from the graphing initialization. Figure \ref{fig:subfigB} provides the objective throughout the entire model fitting procedure. The final partition and hard classifications are provided in Figures \ref{fig:subfigC} and \ref{fig:subfigD}, respectively. We see that the initial number of clusters chosen is ten with the turtle shell correctly reducing that number down to six. Importantly, we see a jump in the objective function with each merge followed by a steady increase in the objective as the parameters adjust to the reduced number of clusters.

\section{Results} \label{results}

\subsection{Simulation Studies}
We use three simulation studies to test the performance of the turtle shell when applied to datasets containing characteristics that are typically problematic for mixture model-based clustering methods. The first simulation, the `cross simulation', tests clustering in the presence of non-convexity. The second simulation, the `mixture of Gaussians simulation', tests clustering when the mathematically correct solution doesn't align with the intuitive solution. Lastly, the third simulation, the `outlier simulation', tests clustering when there are a large number of outliers in the dataset. For all simulations, we compare the full turtle shell and simple turtle shell, using $k=25$ for the graphing initialization, to the EM estimation of a GMM with both the BIC and the ICL as the order selection criterion for $K=2,\ldots,9$. In doing so, evidence for the full turtle shell is exhibited. 

\subsubsection{Cross Simulation}\label{sec:cross}
\begin{figure}[ht!]% 
  \centering
    \subfloat[\centering  Estimated clusters from a GMM estimated by the EM with the BIC and the ICL.]{{\includegraphics[width=0.43\linewidth]{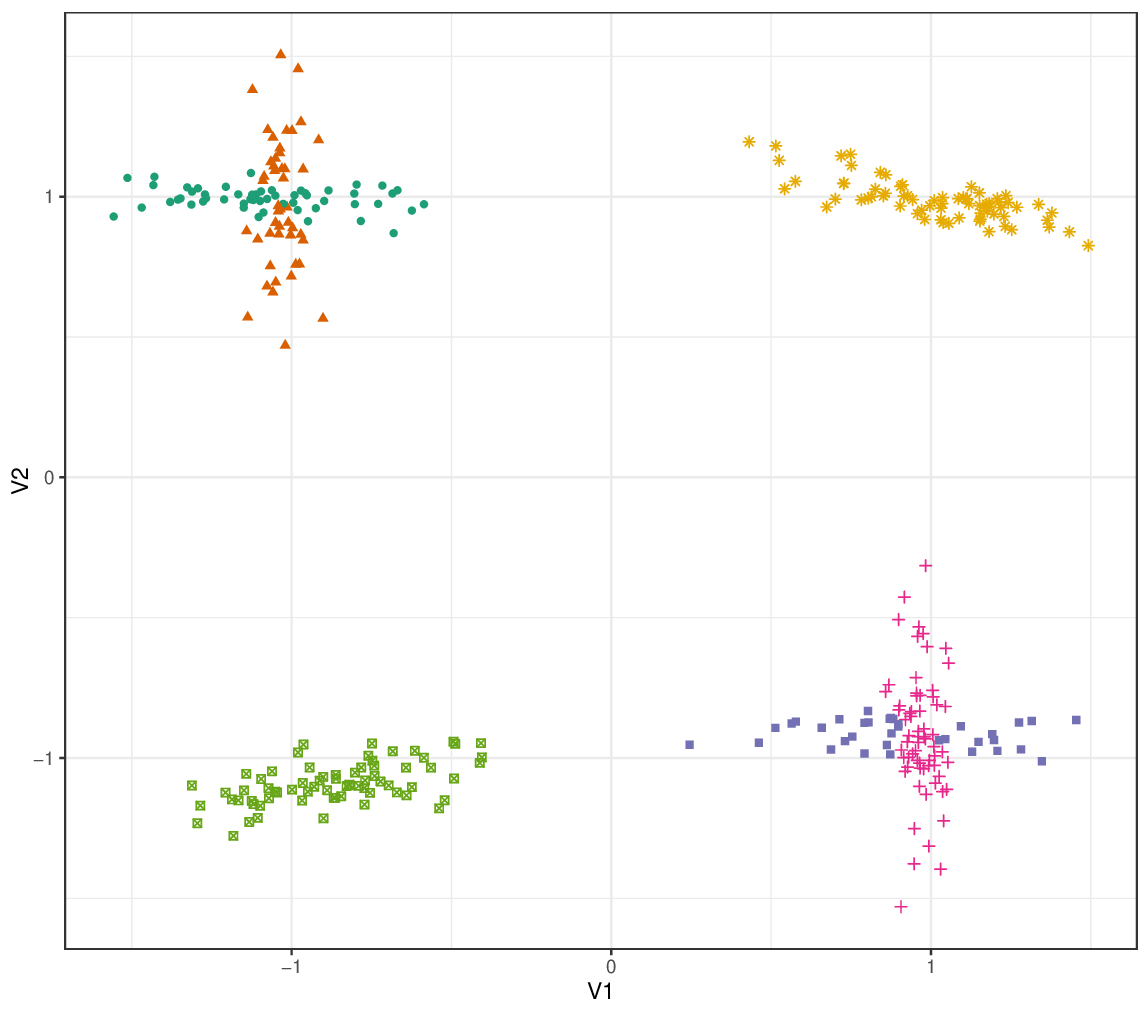}}}%
        \subfloat[\centering  Estimated clusters from the full turtle shell and the simplified turtle shell.]{{\includegraphics[width=0.43\linewidth]{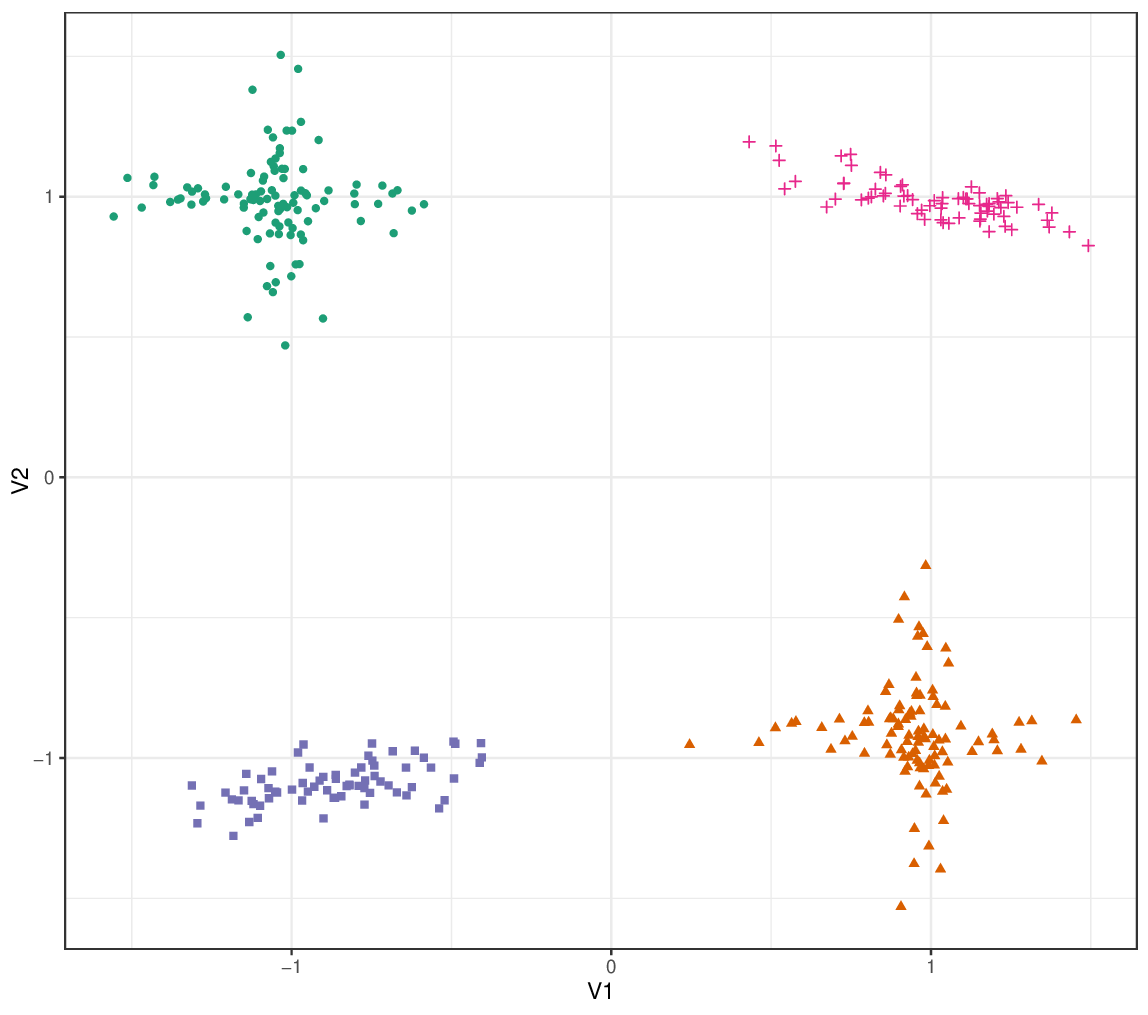}}}%
    \caption{An example result from each tested method on a simulated dataset from the cross simulation.}
    \label{ex:s1}
\end{figure}

This simulation is frequently used to illustrate the performance of clustering methods on non-convex clusters. We simulate data from six multivariate Gaussians; however, we overlap clusters, resulting in two crossed clusters and two Gaussian clusters from the original six Gaussian clusters, visualized in Figure \ref{ex:s1}. We generate 250 similar datasets and summarize the results in Table \ref{tab:cross}.

From Table \ref{tab:cross}, the EM estimation of a GMM with both the BIC and the ICL as the order selection criterion, most often selects six clusters. This is in agreement with the data-generating mechanism, although not fully, as the datapoints at the intersection of the crossed clusters are arbitrarily assigned to one of the two Gaussians. The turtle shell selects four clusters nearly every time, merging the crossed-over Gaussians into crossed clusters. This is true for both the full and simple turtle shell, with the full turtle shell out-performing all methods. 

\begin{table}[ht!]
\caption{Frequency table of the number of clusters selected by each tested method on the cross simulation.}
\centering
\begin{tabular}{|c |c| c| c| c | c | c| c|c|} 
 \hline
  & 2 & 3 & 4 &5&6&7&8 &9 \\ [0.5ex] 
 \hline\hline
Turtle shell  - Full  & 0& 0 & 246 & 2&2&0&0&0\\
 \hline
 Turtle shell  - Simple  & 0&0  & 222 &24 &3&1&0&0\\
 \hline
 GMM EM with the ICL  &0 &0  &41  &22 &136&49&2&0\\ 
\hline
 GMM EM with the BIC & 0& 0 &0  &2 &134&79&31&4\\ 
\hline
\end{tabular}\label{tab:cross}
\end{table}

\subsubsection{Mixture of Gaussians Simulation}\label{sec:mixG}

\begin{figure}[ht!]% 
    \centering
    \subfloat[\centering  Estimated clusters from a GMM estimated by the EM with the BIC.]{{\includegraphics[width=0.43\linewidth]{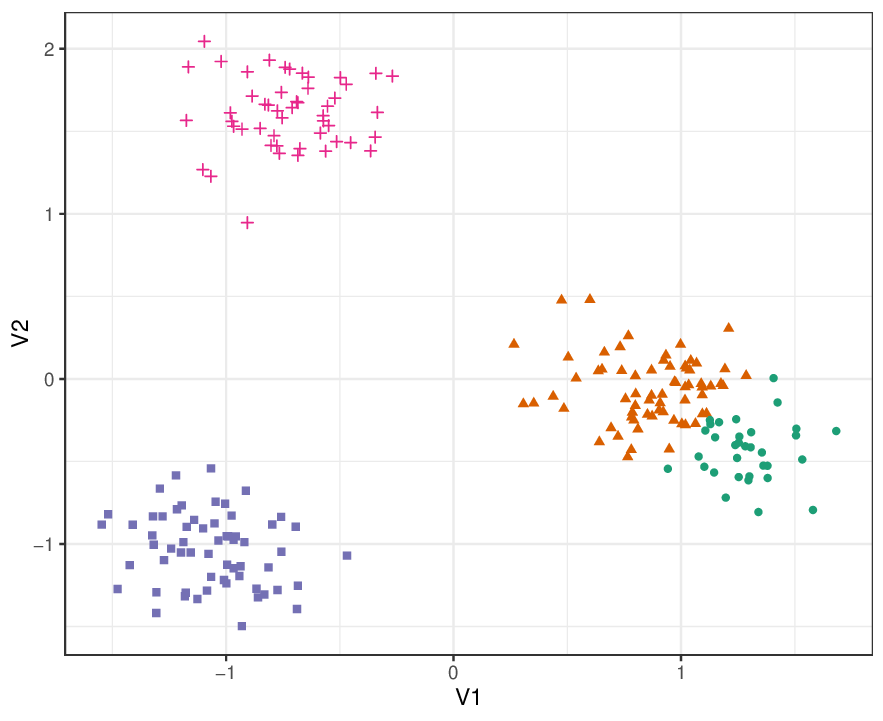}}}%
        \subfloat[\centering  Estimated clusters from a GMM estimated by the EM with the ICL, the full turtle shell, and the simple turtle shell.]{{\includegraphics[width=0.43\linewidth]{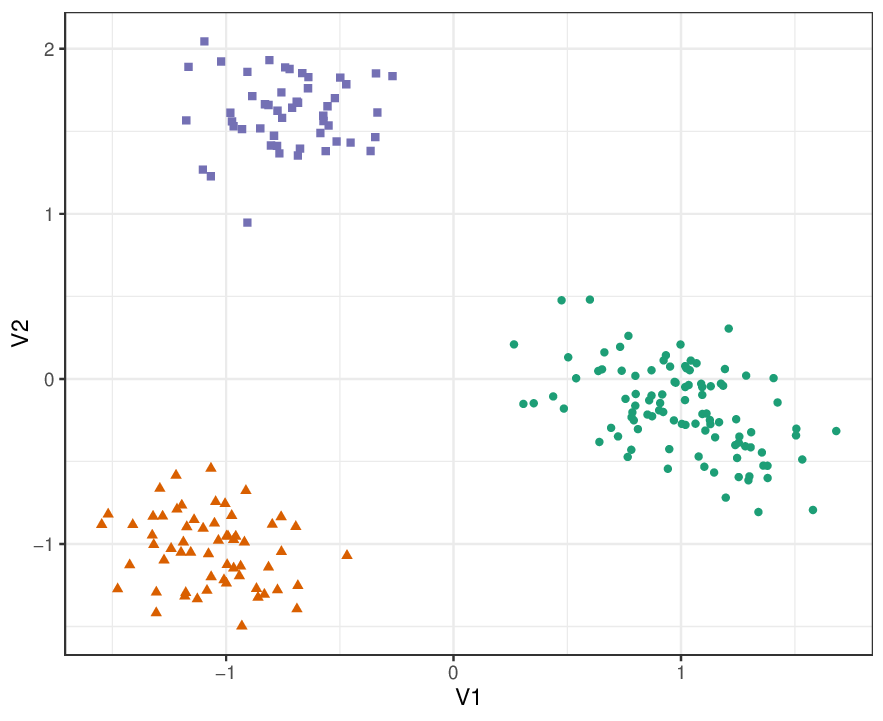}}}%

    \caption{An example result from each tested method on a simulated dataset from the mixture of Gaussians simulation.}
    \label{ex:s2}
\end{figure}

In this simulation, we generate 210 observations from a four-component Gaussian mixture model; however, two of the four clusters are positioned closely, appearing as one cluster. We generate 250 similar datasets and summarize the results in Table \ref{tab:mixG}, with an illustrated example provided in Figure \ref{ex:s2}.

\begin{table}[ht!]
\caption{Frequency table of the number of classes selected by each tested method on the mixture of Gaussians simulation.}
\centering
\begin{tabular}{|c |c| c| c| c | c | c| c|c|} 
 \hline
  & 2 & 3 & 4 &5&6&7&8 &9 \\ [0.5ex] 
 \hline\hline
Turtle shell  - Full &0 & 250 & 0 & 0&0&0&0&0\\
 \hline
 Turtle shell  - Simple & 0& 195 & 53 & 2&0&0&0&0\\
 \hline
 GMM EM with the ICL  &0 & 247 &3  & 0&0&0&0&0\\ 
\hline
 GMM EM with the BIC &0 & 152 &96  & 2&0&0&0&0 \\ 
\hline
\end{tabular}\label{tab:mixG}
\end{table}

From Table \ref{tab:mixG}, the EM estimation of a GMM with the BIC as the order selection criterion often selects the four-component solution rather than the more intuitive three-component solution that is chosen most often by the EM estimation of a GMM with the ICL and the turtle shell. Again, the full turtle shell out-performs all tested methods, including the simple turtle shell.

\subsubsection{Outlier Simulation}\label{sec:out}

\begin{figure}[ht]% 
   \centering
    \qquad
    \subfloat[\centering  Estimated clusters from a GMM estimated by the EM with the BIC.]{{\includegraphics[width=0.43\linewidth]{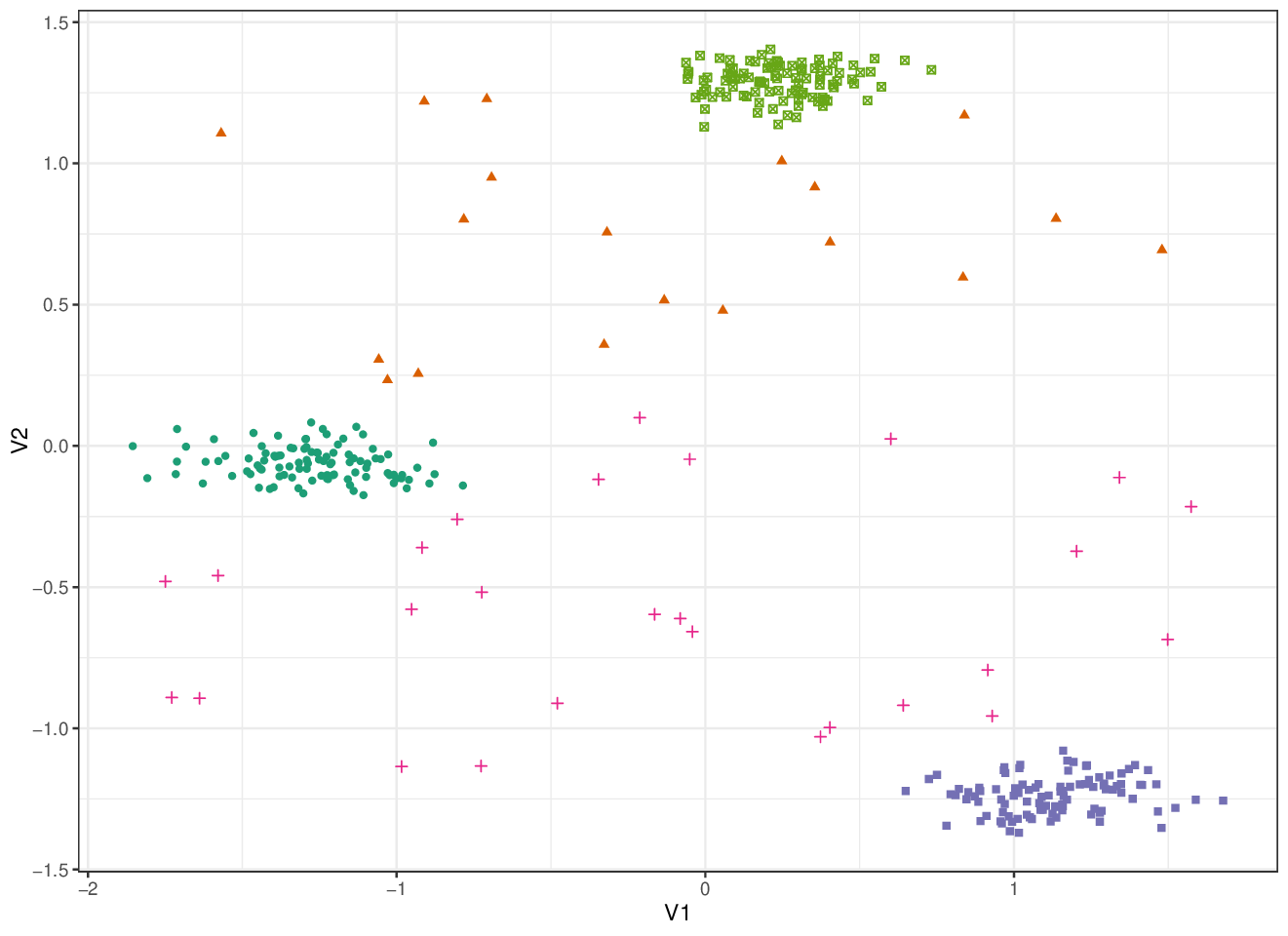}}}%
        \subfloat[\centering  Estimated clusters from a GMM estimated by the EM with the ICL.]{{\includegraphics[width=0.43\linewidth]{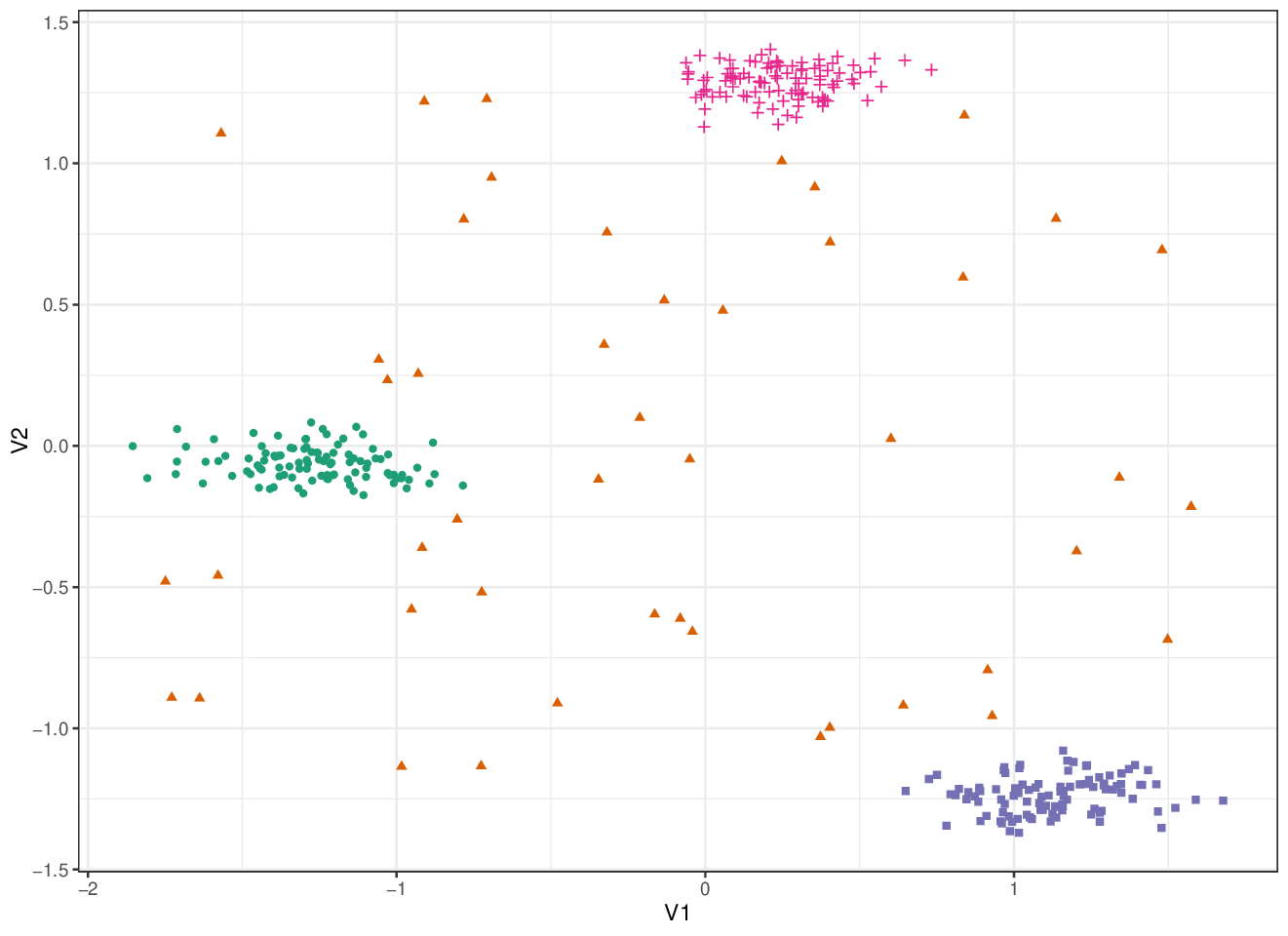}}}%
        
    \centering
     \qquad
        \subfloat[\centering  Estimated clusters from the full turtle shell.]{{\includegraphics[width=0.43\linewidth]{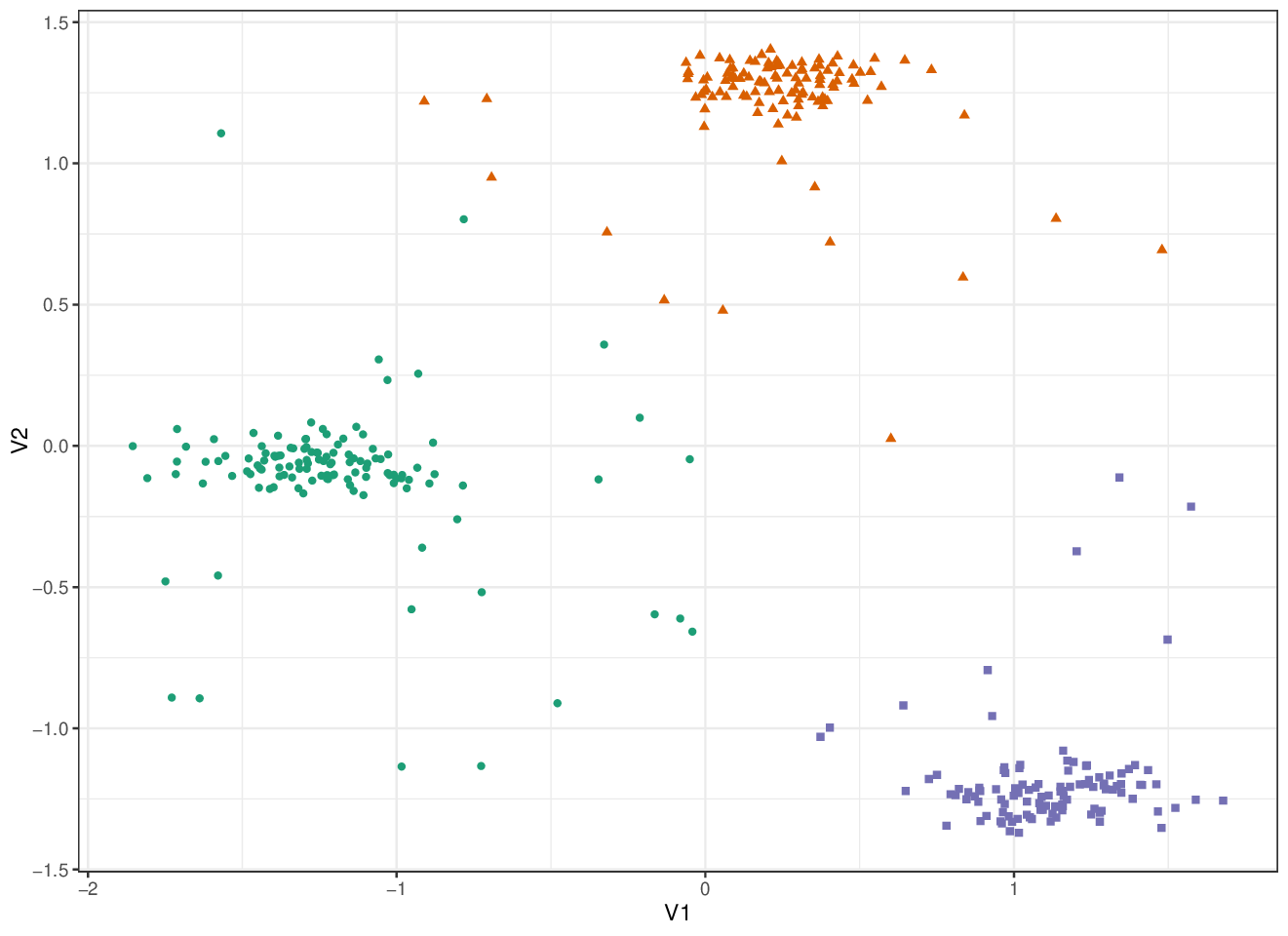}}}%
        \subfloat[\centering  Estimated clusters from the simple turtle shell.]{{\includegraphics[width=0.43\linewidth]{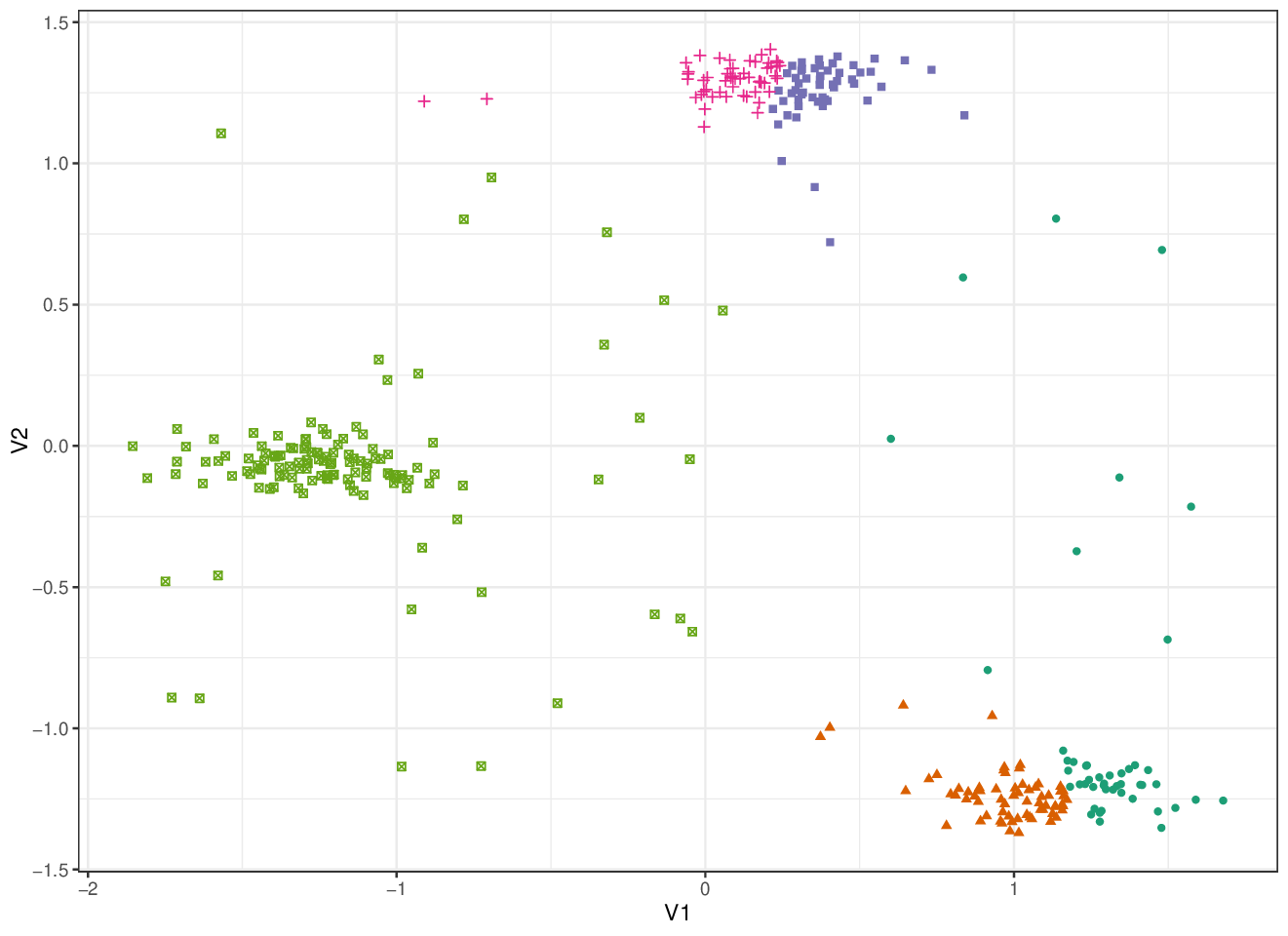}}}
    \caption{An example results from each tested method on a simulated dataset from the outlier simulation.}
    \label{ex:s3}
\end{figure}

In this simulation, we randomly generate 300 observations from a three-component Gaussian mixture model and 50 observations from a uniform distribution over the entire input space. We randomly generate 250 datasets under these conditions and summarize the results in Table \ref{tab:out}, with an illustrated example in Figure \ref{ex:s3}.

Due to the heavy presence of outliers in this simulation, performance varies substantially across the tested methods. For the EM estimation of a GMM with both the BIC and the ICL as the order selection criterion, we see the separation of the outliers into separate outlier clusters. With the full turtle shell, outliers get estimated into the closest Gaussian cluster, whereas the simple turtle shell overestimates the number of clusters by splitting the true Gaussian clusters into smaller unintuitive clusters. 

\begin{table}[ht]
\caption{Frequency of the number of classes selected by each tested method on the outlier simulation.}
\centering
\begin{tabular}{|c |c| c| c| c | c | c| c|c|} 
 \hline
  & 2 & 3 & 4 &5&6&7&8 &9 \\ [0.5ex] 
 \hline\hline
Turtle shell  - Full &0 & 238 & 8 & 1&1&0&0&0\\
 \hline
 Turtle shell  - Simple & 0&58  & 45 &44 &100&1&0&0\\
 \hline
 GMM EM with the ICL  &0 & 0 & 154 &90 &4&0&0&0\\ 
\hline
 GMM EM with the BIC & 0& 0 &139  &109 &0&0&0&0\\ 
\hline
\end{tabular}\label{tab:out}
\end{table}

All three simulations explored herein illustrate the difference between an EM estimation of a mixture model and our proposed method for estimation of a mixture model. Further, these simulations provide evidence for the full turtle shell over the simple turtle shell. As such, we focus solely on the full turtle shell for the real data analyses in Section~\ref{real}.

\subsection{Real Data Applications} \label{real}
In the following real data analyses we continue to compare the full turtle shell to a GMM with the ICL and with the BIC as the order selection criterion. Additionally, we compare our methods to two prominent clustering methods specifically designed for flow cytometry data, as we compare the methods on six standard clustering datasets, four image datasets, and two benchmark flow cytometry datasets. See Table~\ref{datainfo} for further details regarding each data set.

\begin{table}[ht]
\centering
\caption{Information on dimensionality, source, and data type for each tested dataset.}
\label{datainfo}
\begin{tabular}{lccccc}
\hline
  \textbf{Dataset}   & $N$ & $D$& $K$& \textbf{Source} & \textbf{Type of dataset}\\
  \hline
  Bankruptcy&66&2&2&\cite{mixghd21}  &Clustering\\
  Wine&178&13&3&\cite{gclus_pack} & Clustering \\
  Banknote&200&6&2&\cite{mclust_new}& Clustering  \\  
  Seeds&210&7&3&\cite{seeds_236}&  Image \\ 
  Thyroid&215&5&3&\cite{mclust_new}&  Clustering \\     
  Ecoli&336&8&8&\cite{ecoli_39}&  Clustering \\  
 % Vertebral&310&6&3&\cite{vertebral_column_212}&  Image \\  
   \thead[l]{Wholesale \\ Customer}&440&6&2&\cite{wholesale_customers_292}& Clustering \\
   \thead[l]{Breast \\ Cancer} &569&30&2&\cite{breast_cancer}&Image\\
   Pendigits &3498&16&10&\cite{pendigits}&Image\\
      Statlog &4435&36&7&\cite{statlog}&Image\\
   Levine13&81747&13&24&\cite{levine2015data} & Flow Cytometry \\
    HIPC&32948&6&8&\cite{cytoR} & Flow Cytometry \\
   \hline
\end{tabular}
\end{table}

The first flow cytometry method we compare to is cytometree \citep{commenges2018cytometree}. In cytometree, a classification tree is built as follows: for each variable, obtain the AIC values from a one-component univariate GMM and a two-component univariate GMM and calculate the normalized difference in AIC; then, split the variable that maximizes this difference, provided the split is greater than some pre-specified threshold (i.e., $t = 0.1$). This process iterates until no variables remain to split \citep{commenges2018cytometree}.  We include cytometree into the analysis as it was introduced with one of the flow cytometry datasets we explore herein --- the HIPC dataset, available in the \texttt{cytometree} \textsf{R} package \citep{cytoR} --- and thus, we expect this method to perform well on this dataset.

The second flow cytometry method we compare to is PhenoGraph \citep{levine2015data}, as implemented in the \textsf{R} package \texttt{iCellR} \citep{icellr}. PhenoGraph fits a $k$NN using the Jaccard similarity coefficient and then partitions the graph into communities. Parameter $k$, corresponding to the number of neighbours, must be chosen before implementation (i.e., $k=15$). The flow cytometry dataset Levine13 was analyzed in the corresponding manuscript for PhenoGraph \citep{levine2015data}, and therefore, we expect high performance from this method on this dataset. 

Although not included in this analysis, the FlowSOM method proposed by \citet{van2015flowsom} is a prominent clustering method in the analysis of flow cytometry data. In FlowSOM, a self-organizing map (SOM) is built and the nodes are grouped via consensus hierarchical clustering \citep{van2015flowsom}. This method requires some ad-hoc manual labeling of clusters, therefore there is no clear way for us to implement this objectively, as such, it is not included in our analysis.

\begin{figure}[!t]
    \centering
    \includegraphics[width=\linewidth]{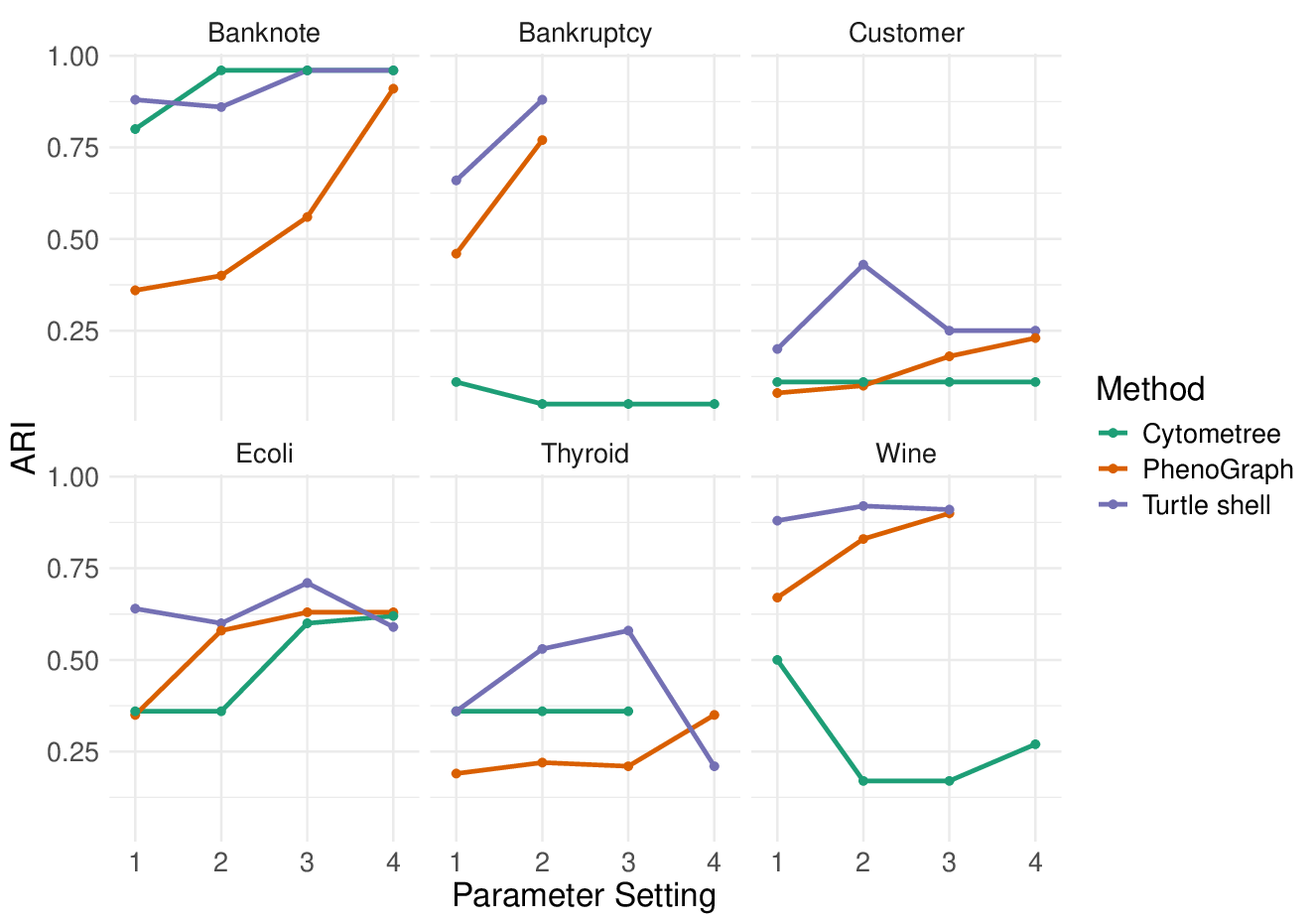}
    \caption{ARI values obtained from each method on each clustering dataset under various parameter settings.}
    \label{k_eff}
\end{figure}

\begin{figure}[!t]
    \centering
    \includegraphics[width=\linewidth]{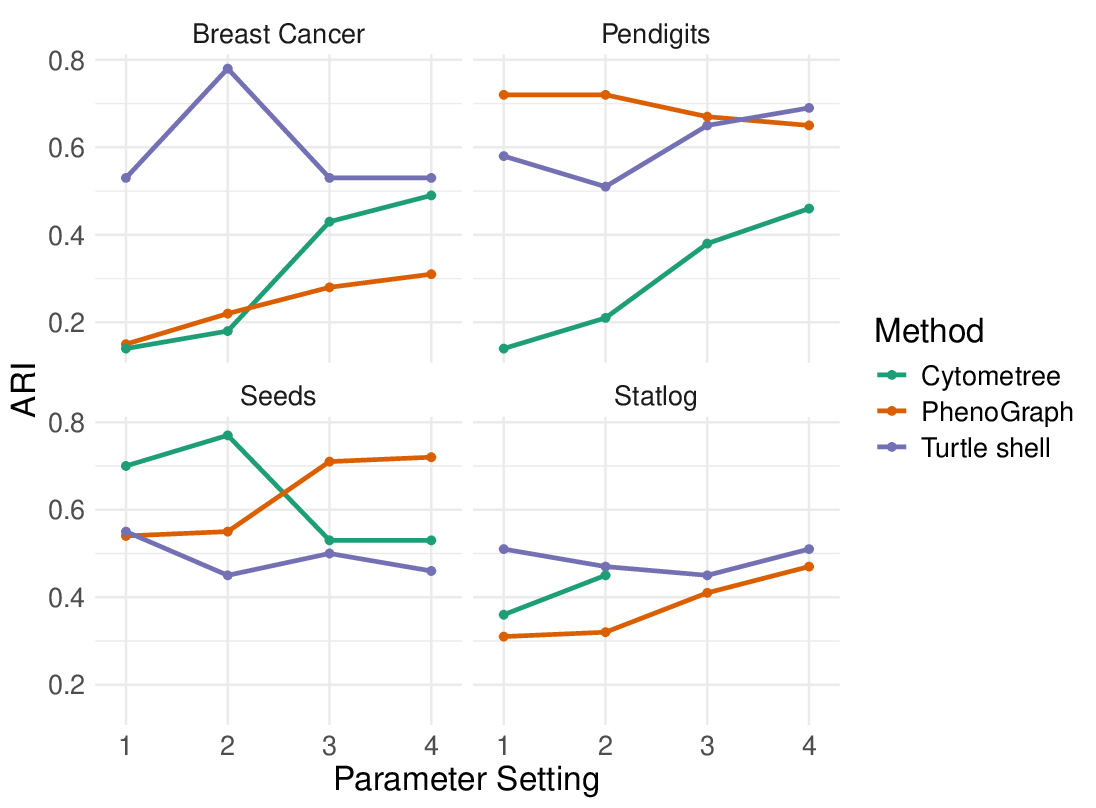}
    \caption{ARI values obtained from each method on each image dataset under various parameter settings.}
    \label{k_eff_image}
\end{figure}

\begin{figure}[!t]
    \centering
    \includegraphics[width=\linewidth]{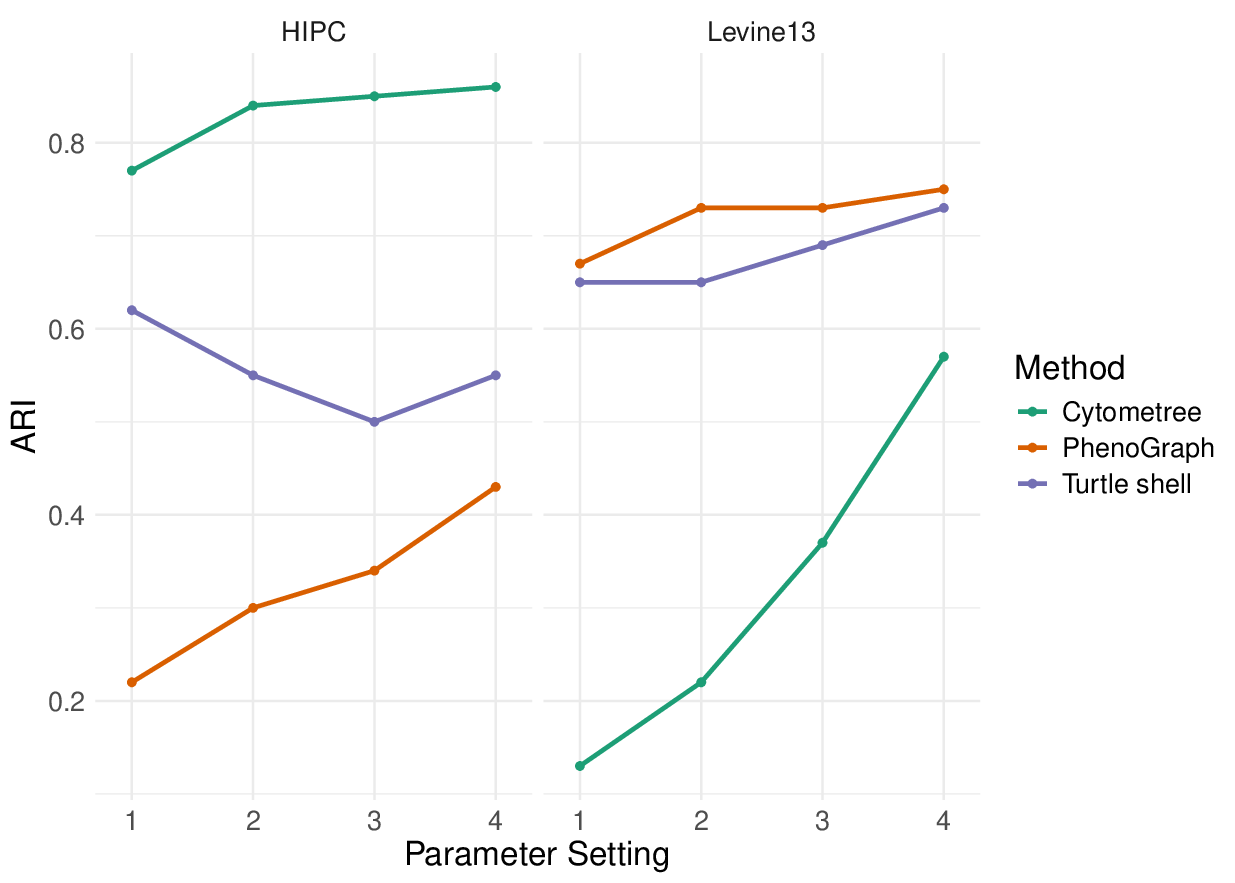}
    \caption{ARI values obtained from each method on each flow cytometry dataset under various parameter settings.}
    \label{k_eff2}
\end{figure}

For all tested methods, if any parameter must be selected \textit{a priori}, we test a range of values based on suggestions in the corresponding manuscript. For PhenoGraph and the turtle shell, we test $k=15, 25, 45, \text{ and }100,$ provided that $k<\sfrac{N}{2}$; however, in practice we would not recommend using $k=100$ for any of the benchmark clustering datasets due to the potential for an underestimation of $K$. For cytometree, we test $t=0.05, 0.1, 0.2, 0.25$ for the normalized AIC threshold. We record the best adjusted Rand index (ARI) \citep{hubert85} and the corresponding estimated number of clusters for all methods in Table \ref{tab:res}. To compare the robustness of each method to initial parameter values see Figures \ref{k_eff}, \ref{k_eff_image}, and \ref{k_eff2}.

%\begin{figure}[ht]
 %   \centering
 %   \includegraphics[scale=0.5]{bankruptcy_true.eps}
 %   \caption{Scatterplot of the true clusters on benchmark clustering dataset bankruptcy.}
 %   \label{bankrupt}
%\end{figure}

\begin{table}[ht]
\centering
\caption{ARI and estimated number of clusters, $(K)$, from each method on benchmark and flow cytometry datasets. The darker grey corresponds to the highest ARI for a given dataset and the lighter grey, the second highest ARI.}
\label{tab:res}
\begin{tabular}{lccccc}
\hline
\hline
  \textbf{Dataset}   & \textbf{Turtle shell} & \textbf{\thead{GMM EM \\ with ICL}}&  \textbf{\thead{GMM EM \\ with BIC}}& \textbf{Cytometree} &   \textbf{PhenoGraph}  \\
\hline
\hline
Bankruptcy & \cellcolor{Gray}0.88 (2) & 0.58 (3)& 0.58 (3) &0.11 (3)  &\cellcolor{lightGray}0.77 (3) \\
Wine & \cellcolor{lightGray}0.92 (3)& \cellcolor{Gray}0.93 (3)&\cellcolor{Gray}0.93 (3) & 0.50 (10) &0.90 (3)\\
Banknote & \cellcolor{Gray}0.96 (2) & \cellcolor{lightGray}0.86 (3) &0.68 (4) &\cellcolor{Gray}0.96 (2)  & 0.91 (2) \\
Seeds &0.55 (4) &0.58 (4) & 0.58 (4)&\cellcolor{Gray}0.77 (3)  &\cellcolor{lightGray}0.72 (3)\\
%Glass & \cellcolor{Gray}0.25 (3) &0.15 (5)&0.15 (5) &  0.12 (7) &\cellcolor{lightGray}0.20 (7)\\
Thyroid &\cellcolor{lightGray}0.58 (3) &\cellcolor{Gray}0.89 (3) & \cellcolor{Gray}0.89 (3) &0.36 (2)  &0.35 (4) \\
Ecoli &\cellcolor{Gray} 0.71 (3) &\cellcolor{lightGray}0.64 (6) & \cellcolor{lightGray}0.64 (6) & 0.62 (8) & 0.63 (3)\\
%Vertebral & \cellcolor{lightGray}0.35 (2) &\cellcolor{Gray}0.40 (3) &\cellcolor{Gray} 0.40 (3)& -0.01 (3) & 0.28 (5)\\
\thead[l]{Wholesale \\ Customer}   & \cellcolor{Gray}0.43 (2) &0.11 (7) & 0.11 (7)  &0.11 (7) & \cellcolor{lightGray}0.23 (4) \\
\thead[l]{Breast \\ Cancer}   & \cellcolor{Gray}0.78 (2) & 0.17 (7) & 0.17 (7) &0.49 (7) & 0.31 (5) \\
Pendigits &\cellcolor{lightGray}0.69 (12) & 0.50 (27) & 0.44 (32) &0.46 (96) & \cellcolor{Gray}0.72 (12) \\
Statlog &\cellcolor{Gray}0.51 (7) & 0.33 (17) & 0.36 (14) &0.45 (108) & \cellcolor{lightGray}0.47 (10) \\
Levine13 &  \cellcolor{lightGray}0.73 (10) & 0.08 (195)&0.08 (195) &  0.57 (190) &\cellcolor{Gray}0.75 (16)\\
HIPC    & \cellcolor{lightGray}0.62 (5) &0.52 (9) & 0.16 (41)&  \cellcolor{Gray}0.86 (9) &0.42 (13)\\
\hline
\hline
\end{tabular}
\end{table}

From Table \ref{tab:res}, the turtle shell obtains the highest or second-highest ARI across nearly all tested datasets, with the only exception being the Seeds dataset. For this dataset, the turtle shell results in two clearly separated groups, rather than the `true' three groups. However, the three-group solution is obtained from some $\lambda_1, \lambda_2$ settings but is not selected by the ASW. When a different metric is used, such as the DBCV index \cite{moulavi2014density}, the three-cluster solution is chosen and an ARI of 0.78 is achieved. This behaviour is unsurprising, as the DBCV index is less sensitive to the irregular cluster shape expected in image data than ASW. For the Thyroid dataset, the turtle shell obtains three clusters that, when visualized, appear to be very similar to the categorical variable we compare to, with boundary lines drawn slightly differently, resulting in an ARI of 0.58. For all other datasets tested, the turtle shell obtains a high ARI and an appropriate number of clusters. This is also true for the flow cytometry datasets, wherein the turtle shell obtains high ARI values on each dataset.  For the flow cytometry datasets, the only method to obtain a higher ARI to the turtle shell is the method introduced with the dataset --- PhenoGraph for Levine13 and cytometree for HIPC. These results reflect the generalizability of the turtle shell to various clustering problems and datasets.

From Fig. \ref{k_eff}, \ref{k_eff_image}, and \ref{k_eff2}, we see the effects of the initial parameter choice on PhenoGraph, cytometree, and the turtle shell. We see fairly stable results from the turtle shell on all datasets across values of $k$. Interestingly, instability in the turtle shell seems to arise from the method providing a result that is either similar to PhenoGraph or to the EM estimation of a GMM, depending on the value of $k$. In practice, we recommend varying $k$ and choosing the final result either based on expert knowledge or some intrinsic metric(s), such as the ASW. 


\section{Discussion \& Future Work}\label{discuss}
An unsupervised, probabilistic, discriminative clustering method, called the turtle shell, has been introduced, wherein the data are assumed to belong to a mixture of mixtures of Gaussian and uniform distributions. In this assumption, a non-linear partitioning grid that is robust to high-density areas and noise is obtained, as demonstrated on various real and simulated examples. Automatic selection of the number of clusters is established, and its effectiveness is demonstrated on various simulated examples. 

Two mixture models were explored for the formulation of the conditional model and we found that the mixture of mixtures of Gaussian and uniform distributions consistently outperformed the simpler mixture of Gaussian distributions; however, future steps could include the exploration of other mixture models. Additionally, we find the turtle shell extends well to difficult clustering problems arising in image analysis and immunological research, and leave exploration of other novel applications as an open area of research. 

\addcontentsline{toc}{section}{Bibliography}

\bibliographystyle{chicago}

\bibliography{mcnicholas2022.bib}

\newpage

%\begin{appendices}

\appendix
\section{Appendix: Gradients}\label{appendix:appA}
\noindent
For clarity, define $p_{ki}\equiv p(y_k|\vecx_i;\vecvartheta_k)$, $\hat{p}_{k}\equiv \hat{p}(y_k;\vecvartheta_k)$, $\mathcal{N}_{k}\equiv \mathcal{N}(\vecmu_k,[\matL_k\matL_k^\top]^{-1})$, and $U_k \equiv U(\veca_k,\vecb_k)$.
\subsection{Gradient of $\vecmu_k$}
\noindent
The gradient of the objective function with respect to $\vecmu_k$ is determined by 
\begin{align}\label{grad_mu}
\frac{\partial F}{\partial \vecmu_{k}} & = \frac{\partial}{\partial \vecmu_{k}} (I-  R_1 - R_2),\nonumber \\
&= \frac{1}{N} \sum_{i=1}^N \sum_{c=1}^K \frac{\partial p_{ci}}{\partial \vecmu_{k}} \text{log}\frac{p_{ci}}{\hat{p}_{c}} -\frac{\partial R_1}{\partial \vecmu_{k}}- \frac{\partial R_2}{\partial \vecmu_{k}}.
\end{align}
\subsubsection{Conditional model:}\label{cond_mu}
\noindent
To obtain the gradient of the objective function with respect to $\vecmu_k$, we first obtain the gradient of the conditional model, $p_{ci}$, as 

\noindent
if $c=k,$
\begin{align*}
\frac{\partial}{\partial \vecmu_{k}} p_{ci}=  A_c\frac{ s(\pi_c) \omega_c \mathcal{N}_{c}}{\sum_{g=1}^K  s(\pi_g) [\omega_g\mathcal{N}_{g}+ (1- \omega_g)U_g]} - A_kp_{ci}\frac{s(\pi_k) \omega_k \mathcal{N}_{k}}{\sum_{g=1}^K  s(\pi_g) [\omega_g\mathcal{N}_{g} + (1- \omega_g)U_g]},
\end{align*}
%\vspace{1em}

\noindent
if $c \neq k$,
\begin{align*}
\frac{\partial}{\partial \vecmu_{k}} p_{ci} &= -   A_k p_{ci} \frac{ s(\pi_k) \omega_k \mathcal{N}_{k}}{\sum_{g=1}^K  s(\pi_g) [\omega_g\mathcal{N}_{g} + (1- \omega_g)U_g]},
\end{align*}
where $A_k=(\matL_k\matL_k^\top)(\vecx - \vecmu_k)$.

\subsubsection{Regularizing terms:}\label{reg_mu}
\noindent
Next the gradient of both regularization terms with respect to $\vecmu_k$ is obtained as
\begin{align*}
\frac{\partial R_1}{\partial \vecmu_{k}} &=0,\\
\frac{\partial R_2}{\partial \vecmu_{k}} &=-2(\matL_k\matL_k^\top) \bigg(\frac{\vecb_k+\veca_k}{2} - \vecmu_k\bigg).
\end{align*}
\subsubsection{Final gradient:}
\noindent
By substituting the expressions from Sections \ref{cond_mu} and \ref{reg_mu} into Eq. \ref{grad_mu}, we obtain the final gradient as
\begin{align*}
\begin{split}
\frac{\partial F}{\partial \vecmu_{k}} &= \frac{1}{N} \sum_{i=1}^N  (\matL_k\matL_k^\top)(\vecx - \vecmu_k)\frac{s(\pi_k) \omega_k \mathcal{N}_k}{\sum_{g=1}^K s(\pi_g) [\omega_g\mathcal{N}_g+ (1- \omega_g)U_g]}  \bigg( \text{log}\frac{p_{ki}}{\hat{p}_k}- \sum_{c=1}^K p_{ci} \text{log} \frac{p_{ci}}{\hat{p}_c} \bigg) \\ & \hspace{250pt} +2\lambda_2(\matL_k\matL_k^\top)\bigg(\frac{\vecb_k+\veca_k}{2} - \vecmu_k\bigg).
\end{split}
\end{align*}
\subsection{Remaining gradients}
\noindent
All remaining gradients are obtained similarly and provided below.
\begin{align*}
\begin{split}
\frac{\partial F}{\partial \matL_{k}} &= \frac{1}{N} \sum_{i=1}^N  [-(\vecx-\vecmu_k)(\vecx-\vecmu_k)^\top\matL_k + \matL_k^{-\top}]\frac{s(\pi_k) \omega_k \mathcal{N}_k}{\sum_{g=1}^K s(\pi_g) [\omega_g\mathcal{N}_g+ (1- \omega_g)U_g]}  \bigg( \text{log}\frac{p_{ki}}{\hat{p}_k}- \sum_{c=1}^K p_{ci} \text{log} \frac{p_{ci}}{\hat{p}_c} \bigg) \\& \hspace{250pt} -2\lambda_2 \bigg(\frac{\vecb_k+\veca_k}{2} - \vecmu_k\bigg)\bigg(\frac{\vecb_k+\veca_k}{2} - \vecmu_k\bigg)^\top \matL_k
\end{split}
\\[2ex]
\frac{\partial F}{\partial \pi_{k}} &= \frac{1}{N} \sum_{i=1}^N  [1-s(\pi_k)] p_{ki} \bigg( \text{log}\frac{p_{ki}}{\hat{p}_k}- \sum_{c=1}^K p_{ci} \text{log} \frac{p_{ci}}{\hat{p}_c} \bigg)  +\lambda_1 [1-K s(\pi_k)] \\[2ex]
\frac{\partial F}{\partial \veca_{k}} &= \begin{bmatrix} \frac{\partial I}{\partial a_{k1}} \\ \vdots \\ \frac{\partial I}{\partial a_{kD}} \end{bmatrix} -\lambda_2(\matL_k\matL_k^\top)\bigg(\frac{\vecb_k+\veca_k}{2} - \vecmu_k\bigg) \\[2ex]
\frac{\partial I}{\partial a_{kj}} &= \frac{1}{N} \sum_{i=1}^N \bigg[(b_{kj} - a_{kj})^2 \prod_{d=1, d\neq j}^D (b_{kd} - a_{kd})\bigg]^{-1}  \frac{s(\pi_k)(1- \omega_k)}{\sum_{g=1}^K s(\pi_g) [\omega_g\mathcal{N}_g+ (1- \omega_g)U_g]}   \bigg( \text{log}\frac{p_{ki}}{\hat{p}_k}- \sum_{c=1}^K p_{ci} \text{log} \frac{p_{ci}}{\hat{p}_c} \bigg) \\[2ex]
\frac{\partial F}{\partial \vecb_{k}} &= \begin{bmatrix} \frac{\partial I}{\partial b_{k1}} \\ \vdots \\ \frac{\partial I}{\partial b_{kD}} \end{bmatrix} -\lambda_2(\matL_k\matL_k^\top) \bigg(\frac{\vecb_k+\veca_k}{2} - \vecmu_k\bigg) \\[2ex]
\frac{\partial I}{\partial b_{kj}} &= \frac{-1}{N} \sum_{i=1}^N \bigg[(b_{kj} - a_{kj})^2 \prod_{d=1, d\neq j}^D (b_{kd} - a_{kd})\bigg]^{-1} \frac{s(\pi_k)(1- \omega_k)}{\sum_{g=1}^K s(\pi_g) [\omega_g\mathcal{N}_g+ (1- \omega_g)U_g]}  \bigg( \text{log}\frac{p_{ki}}{\hat{p}_k}- \sum_{c=1}^K p_{ci} \text{log} \frac{p_{ci}}{\hat{p}_c} \bigg) \\[2ex]
\frac{\partial F}{\partial \omega_{k}} &= \frac{1}{N} \sum_{i=1}^N \frac{s(\pi_k)(\mathcal{N}_k - U_k)}{\sum_{g=1}^K s(\pi_g) [\omega_g\mathcal{N}_g+ (1- \omega_g)U_g]}  \bigg( \text{log}\frac{p_{ki}}{\hat{p}_k}- \sum_{c=1}^K p_{ci} \text{log} \frac{p_{ci}}{\hat{p}_c} \bigg) \\[2ex]
\end{align*}
\end{document}